\pdfoutput=1
\documentclass[11pt]{article}
\usepackage[table,xcdraw,dvipsnames]{xcolor}
\usepackage[]{acl}
\usepackage{times}
\usepackage{latexsym}
\usepackage[T1]{fontenc}
\usepackage[utf8]{inputenc}
\usepackage{microtype}
\usepackage{arydshln} 
\usepackage{inconsolata}
\usepackage{graphicx}
\usepackage{subcaption}
\usepackage{setspace}
\usepackage{amsmath,amsthm,amsfonts,amssymb,bm}
\usepackage{hyperref}
\usepackage{url}
\usepackage{adjustbox} 
\usepackage{listings}
\usepackage{multirow}
\usepackage{caption}
\usepackage{colortbl}
\usepackage[most]{tcolorbox}
\usepackage{titletoc}
\usepackage[capitalize]{cleveref} 
\usepackage{float}
\usepackage{tabularx}
\usepackage{booktabs}
\usepackage{array}
\usepackage{makecell}
\usepackage{enumitem}
\usepackage{bbold}
\usepackage[misc]{ifsym}
\usepackage{CJKutf8}
\usepackage{CJK}
\usepackage{color}
\usepackage{bbding}
\usepackage[normalem]{ulem}
\useunder{\uline}{\ul}{}

\definecolor{mygreen}{HTML}{D5E8D4}
\definecolor{orange}{HTML}{FFE6CC}

\definecolor{lightblue}{HTML}{DAE8FC}
\definecolor{red}{RGB}{255, 0, 0}
\definecolor{green}{RGB}{120, 220, 130}
\definecolor{lightgreen}{RGB}{193, 255, 193}

\newcommand{\mycolorbox}[4]{
  \begin{tcolorbox}[
    breakable, 
    title={\textcolor{black}{\textbf{#1}}}, 
    colframe={#3}, 
    colback={#4},   
    width=\textwidth
  ]
  #2\\
  \end{tcolorbox}
}

\usepackage{afterpage}

\title{CPO: Addressing Reward Ambiguity in Role-playing Dialogue via Comparative Policy Optimization}

\author{
    Xinge Ye$^{1,2}$, 
    Rui Wang$^{1}$,
    Yuchuan Wu$^{1}$,
    Victor Ma$^{3}$,
    Feiteng Fang$^{1}$,
    Fei Huang$^{1}$,
    Yongbin Li$^{1}$\Thanks{ Corresponding Author}\\
    \footnotesize${}^1$Tongyi Lab, Alibaba Group; \footnotesize${}^3$Independent Researcher \\
    \footnotesize${}^2$School of Artificial Intelligence, University of Chinese Academy of Sciences, Beijing, China\\ 
    \footnotesize{xinge.ye555@gmail.com}; \footnotesize{\{wr224079, shuide.lyb\}@alibaba-inc.com} \\
}

\begin{document}
\maketitle
\begin{abstract}
Reinforcement Learning Fine-Tuning (RLFT) has achieved notable success in tasks with objectively verifiable answers (e.g., code generation, mathematical reasoning), yet struggles with open-ended subjective tasks like role-playing dialogue. Traditional reward modeling approaches, which rely on independent sample-wise scoring, face dual challenges: subjective evaluation criteria and unstable reward signals.
Motivated by the insight that human evaluation inherently combines explicit criteria with implicit comparative judgments, we propose \textbf{Comparative Policy Optimization (CPO)}. CPO redefines the reward evaluation paradigm by shifting from sample-wise scoring to comparative group-wise scoring.
Building on the same principle, we introduce the \textbf{CharacterArena} evaluation framework, which comprises two stages:
(1) \textit{Contextualized Multi-turn Role-playing Simulation}, and (2) \textit{Trajectory-level Comparative Evaluation}. By operationalizing subjective scoring via objective trajectory comparisons, CharacterArena minimizes contextual bias and enables more robust and fair performance evaluation.
Empirical results on CharacterEval, CharacterBench, and CharacterArena confirm that CPO effectively mitigates reward ambiguity and leads to substantial improvements in dialogue quality.
 
\end{abstract}

\section{Introduction}

\begin{figure}[th!]
    \centering
    \includegraphics[width=\linewidth]{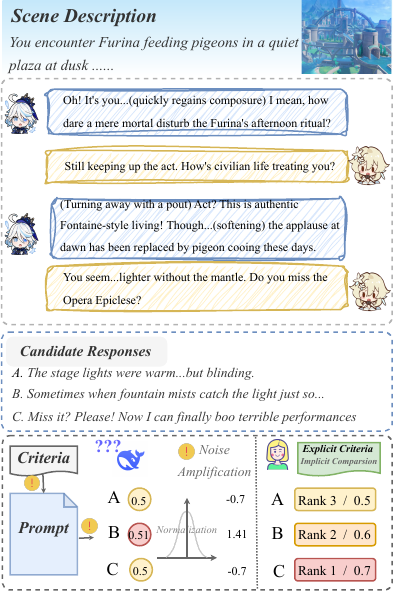}
    \caption{Challenges in role-playing reward estimation. Sample-wise LLM evaluation suffers from ambiguous criteria, prompt sensitivity, and scoring instability, amplifying errors in GRPO's advantage computation. Group-wise rewarding provides clearer evaluation criteria and implicit comparisons. Additionally, ranking-based evaluation is simpler and more stable than independent scoring, reducing error propagation in advantage estimation.}
    \label{fig:intro}
\end{figure}

Role-playing dialogue systems aim to support immersive multi-turn interactions by simulating specific character personas \cite{characterbench, RoleLLM}. A core challenge lies in generating responses that are not only coherent with the character profile but also rich in narrative appeal and stylistic diversity \cite{zhou-etal-2024-characterglm}. Existing approaches primarily adopt Supervised Fine-Tuning (SFT) \cite{sft} over high-quality dialogue corpora \cite{wang-etal-2024-rolellm, CoSER}, following an imitation learning paradigm. While SFT has proven effective to a degree, it is inherently constrained by the distribution of the training data—leading to overfitting on observed dialogue patterns, limited creative flexibility, and often underwhelming performance in delivering truly engaging and immersive role-playing experiences \cite{sun2025inverserlignmentlargelanguagemodel}.

Recent advances in Reinforcement Learning Fine-Tuning (RLFT) \cite{rlhf, rlaif, dpo} present a promising alternative. RLFT has achieved notable success as a post-training strategy for large language models (LLMs), particularly in aligning outputs with human preferences \cite{gpt-4} and enhancing complex reasoning abilities \cite{deepseekr1, qwen2.5, yang2025qwen3technicalreport}. 

However, the effectiveness of RLFT hinges on the availability of reliable and discriminative reward signals — a condition that holds in well-structured domains with objective correctness criteria \cite{liu2025inferencetimescalinggeneralistreward}. In stark contrast, role-playing poses a fundamentally more subjective and open-ended challenge, where key objectives like personality consistency, narrative appeal, and emotional resonance are inherently ill-defined. As illustrated in Figure~\ref{fig:intro}, current reward modeling methods encounter two limitations.\textit{ First,} the evaluation criteria for open-ended responses are intrinsically ambiguous, making it difficult to establish reliable and consistent scoring rules. \textit{Second,} existing sample-wise LLM-based evaluators are highly sensitive to prompt variations, often producing unstable and weakly discriminative scores, and in some cases, collapsing most outputs into a narrow scoring range \cite{batcheval}.

To address these issues and unlock the potential of RLFT in role-playing, we propose \textbf{Comparative Policy Optimization (CPO)}. Inspired by the observation that human evaluation relies not only on explicit criteria but also on implicit comparisons between samples \cite{batcheval}, CPO reframes reward modeling from individual sample scoring to group-wise scoring. This shift leads to an approximately 20\% improvement in human agreement, effectively reducing reward ambiguity. Furthermore, we introduce \textbf{CharacterArena}, a comprehensive evaluation framework. It comprises two stages: \textbf{(1) Contextualized Multi-turn Role-playing Simulation}, which generates interaction trajectories under controlled character and scenario settings; and \textbf{(2) Trajectory-based Comparative Evaluation}, which enables fair and robust assessments by anchoring evaluations to direct trajectory comparisons rather than absolute scores. Experimental results demonstrate that CPO consistently outperforms other RLFT methods on both CharacterEval and CharacterBench, and surpasses all baselines within the CharacterArena evaluation framework. These findings highlight the effectiveness of our approach in addressing reward ambiguity and substantially enhancing dialogue quality in role-playing systems.

Our main contributions are as follows:
\begin{itemize}[itemsep= 0.1pt,topsep = 0.1pt,partopsep=0.1pt]
\item We propose Comparative Policy Optimization (CPO), a new RLFT method based on a group-wise reward modeling paradigm, specifically designed to reduce reward ambiguity in open-ended role-playing dialogues.
\item We present CharacterArena, a new evaluation framework that transforms subjective judgments into more objective comparisons, effectively minimizing contextual bias and enabling a fairer assessment of role-playing performance.
\item Experimental results on three benchmarks show that CPO outperforms existing RLHF methods. Further analysis demonstrates that group-wise scoring improves human agreement by up to 20\%.
\end{itemize}
\section{Related Work}
\subsection{Role-playing}
The development of Role-Playing Agents has been revolutionized by recent breakthroughs in LLMs \cite{llama3, qwen2.5, gpt-4}, empowering users with unprecedented freedom to create and customize their own characters for engaging interactions\cite{CAMEL, chen-etal-2023-large, wang-etal-2024-rolellm, Lu0ZZ24,chen2024personapersonalizationsurveyroleplaying}. This customization often relies on two approaches: prompting general-purpose LLMs for role-play \cite{characterbox}, or developing specialized LLMs specifically tailored for character generation by training them on role-playing dialogues \citet{zhou-etal-2024-characterglm, characterdestiny, CoSER}. 

Current evaluation of Role-Playing Agents (RPAs) employs two primary paradigms: question-answering (QA) and LLM-based judging benchmarks. QA benchmarks typically use multiple-choice questions to target specific RPA capabilities, such as character knowledge \cite{shen2024roleevalbilingualroleevaluation}, decision-making \cite{characterdestiny}, motivation recognition \cite{yuan-etal-2024-evaluating}, and personality fidelity \cite{shao-etal-2023-character, incharacter}. Conversely, the LLM-based judging approach prompts RPAs with predefined questions to assess role-playing performance, which is then scored by LLM judges or reward models \cite{tu-etal-2024-charactereval, characterbench, characterbox, MMRole}. This method generally allows for a more comprehensive evaluation, assessing conversational skills, character adherence (knowledge and personality), and interaction engagement. A significant limitation of LLM-judging, however, is its dependence on static, externally provided conversation histories. This overlooks the vital multi-turn dynamics of interactive role-playing and risks introducing biases from context not generated by the agent under evaluation \cite{CoSER}.

\subsection{Reinforcement Learning Fine-tuning}
Reinforcement Learning Fine-tuning (RLFT) has become a widely adopted approach for post-training LLMs at scale, significantly enhancing their emergent capabilities \cite{rft,dpo,rlhf}. A key challenge of RL is to obtain accurate reward signals for LLMs in specific domains. Recent studies on RLFT primarily focus on highly structured tasks with well-defined rules and verifiable ground-truth answers, such as code generation and mathematical reasoning~\cite{yeo2025demystifyinglongchainofthoughtreasoning, tinyzero, zeng2025simplerl, prime}. For example, DeepSeek-R1~\cite{deepseekr1} demonstrates the effectiveness of RLFT using purely rule-based reward functions. 

However, real-world applications often involve more general and complex tasks that lack clear evaluation criteria or deterministic rules~\cite{liu2025inferencetimescalinggeneralistreward, su2025crossingrewardbridgeexpanding}. In this work, we focus on the role-playing setting—a particularly subjective and open-ended domain—to address the ambiguous rewards challenge.

\begin{figure*}[t]
    \centering
    \includegraphics[width=1\linewidth]{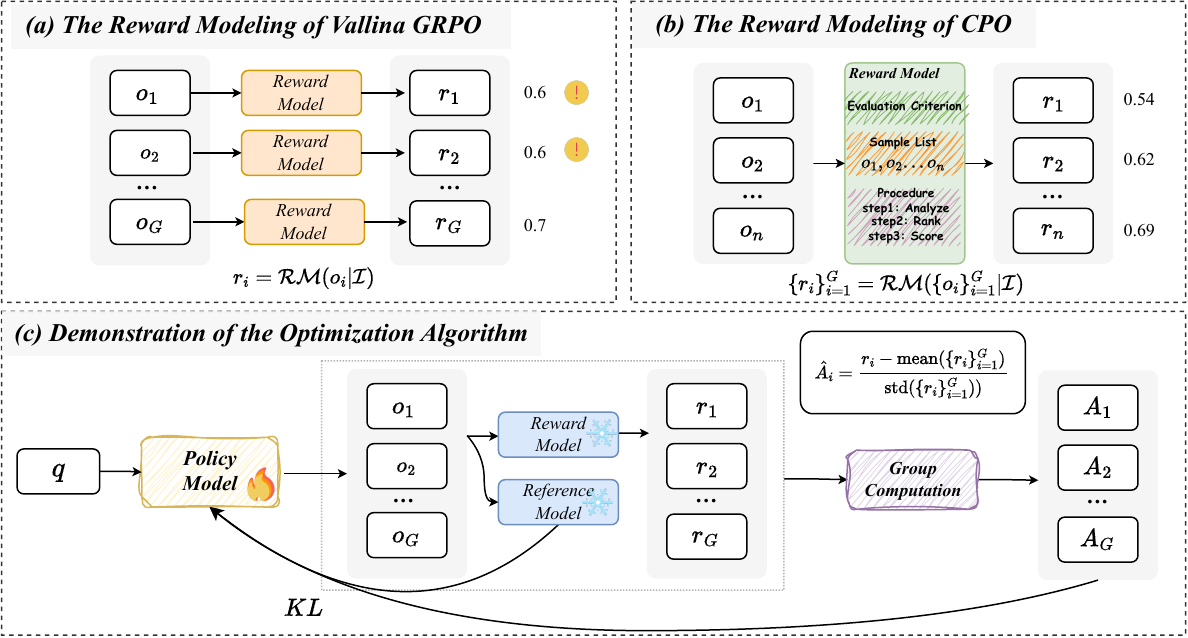}
    \caption{Comparison of GRPO and CPO. GRPO suffers from sample-wise reward ambiguity in role-playing scenarios. CPO addresses this limitation by employing a group-wise approach, comparing responses within groups to establish relative quality. This comparative assessment results in more accurate and stable reward signals than GRPO.
    }
    \label{fig:method}
\end{figure*}

\section{Preliminary}
\label{sec:preliminary}
This section introduces the Group Relative Policy Optimization (GRPO) algorithm \cite{DeepSeekMath}, a reinforcement learning method commonly employed in the RL fine-tuning (RLFT) stage of LLMs. GRPO optimizes the policy by leveraging relative reward comparisons among a group of generated responses.

Formally, let $q$ denote the input query, and $o$ represent the response generated by a policy. We denote the current and old policy models as $\pi_{\theta}$ and $\pi_{\theta_{\text{old}}}$, respectively. For a given query $q$, GRPO samples a set of $G$ responses, $\{o_1, o_2, \dots, o_G\}$, from the old policy $\pi_{\theta_{\text{old}}}$. A reward model $\mathcal{RM}$ then assigns a scalar reward to each response, resulting in $G$ corresponding rewards: 
\begin{equation}
\label{eq:sample-wise-reward}
r_i = \mathcal{RM}(o_i\,|\, \mathcal{I})
\end{equation}
The GRPO objective is then defined as:
\begin{equation}
\label{eq:grpo_1} 
\resizebox{\linewidth}{!}{$
\begin{aligned}
\mathcal{J}_{\text{GRPO}}(\theta) &= \mathbb{E}_{q, \{o_i\}} \Biggl[ \frac{1}{G} \sum_{i=1}^G  \biggl\{ \underbrace{\frac{1}{|o_i|} \sum_{t=1}^{|o_i|} L_{\text{clip}}(o_{i,t})}_{\text{Policy Loss}} - \underbrace{\beta D_{\text{KL}}(\pi_{\theta} || \pi_{\text{ref}})}_{\text{KL Penalty}} \biggr\} \Biggr] 
\end{aligned}
$}
\end{equation}
\begin{equation}
\label{eq:grpo_2}
\resizebox{\linewidth}{!}{$
\begin{aligned}
L_{\text{clip}}=&\min\left( r_{i,t}(\theta) \hat{A}_{i,t}, \right. \left.\text{clip}\left( r_{i,t}(\theta), 1 - \epsilon_{\text{low}}, 1 + \epsilon_{\text{high}} \right) \hat{A}_{i,t} \right) 
\end{aligned}
$}
\end{equation}
where
\begin{equation}
\label{eq:rit}
\begin{aligned}
r_{i,t}(\theta) = \frac{\pi_\theta(o_{i,t} \mid q, o_{i,<t})}{\pi_{\theta_{\text{old}}}(o_{i,t} \mid q, o_{i,<t})}
\end{aligned}
\end{equation}
$\epsilon$ and $\beta$ control the clipping range and the strength of the KL divergence penalty, respectively. $\hat{A}_{i,t}$ represents the advantage estimate, indicating how much better the current response $o_i$ is compared to the average response in the sampled set. It is calculated based on the relative rewards within each group of responses:
\begin{equation}
\label{eq:advantage}
\hat{A}_{i,t} = \frac{r_i-\text{mean}(\{r_i\}_{i=1}^G)}{\text{std}(\{r_i\}_{i=1}^G))}
\end{equation}

\section{Method}

\subsection{Role-playing Task Definition}
\label{sec:role_play}
Role-playing in language agents involves generating dialogue responses that authentically embody a specified character. This task extends beyond conventional open-ended dialogue by requiring the agent to maintain consistent persona traits, align with the character's established background (fictional or real-world), and produce responses that reflect nuanced emotional tones, narrative logic, and appropriate interpersonal dynamics. Formally, given a dialogue history $q=\{u_1, a_1, \cdots, u_t\}$, where $u_i$ and $a_i$ denote user and agent turns, respectively, the model's objective is to generate a response $o \sim \pi( \cdot | q, \mathcal{P})$, sampled from a policy conditioned on both the dialogue history and a predefined character profile $\mathcal{P}$. 

\subsection{Challenges in Role-playing Reward Modeling}
\label{sec:challenges}
While RLFT demonstrates strong performance in objective-driven tasks, its application to role-playing scenarios is fundamentally constrained. As illustrated in Figure \ref{fig:human_evaluation}, traditional sample-wise reward modeling proves inadequate in several respects:

\paragraph{{Ambiguity in Benchmarking:}} Role-playing responses often differ subtly in tone, intent, or persona alignment. The inherent ambiguity in what constitutes a “better” response makes evaluation highly context-dependent. For example, two different responses may receive the same scores due to vague or underspecified evaluation criteria. 

\paragraph{{Scoring Instability:}} When responses are evaluated in isolation, reward scores become highly sensitive to the prompt phrasing and the stochastic nature of LLMs. A response may receive a score of 0.8 on one evaluation, but 0.7 or 0.9 on others. This instability undermines the consistency of rankings and weakens the learning signal derived from them.

\paragraph{{Error Amplification in Advantage Estimation:}} Due to the normalization in advantage estimation, small scoring noise can lead to disproportionately large errors in relative rankings. This issue is especially pronounced when similar responses fall within narrow score intervals, where even minor noise gets magnified after normalization.

\subsection{Comparative Policy Optimization}
\label{sec:cpo}
Human evaluators analyze samples based on the evaluation criterion and provide discriminative scores through comparison between samples \cite{batcheval}. Inspired by this process, we propose \textbf{Comparative Policy Optimization (CPO)}, which mimics human comparative assessment by establishing relative quality benchmarks within response groups. 

As illustrated in Figure~\ref{fig:method}, given evaluation criterion $\mathcal{I}$ and a group of responses $\{o_1, o_2, \dots, o_G\}$ sampled from the old policy $\pi_{\theta_{\text{old}}}$, we define the group-comparative reward assignment as:
\begin{equation}
\label{eq:group-wise-reward}
\{r_i\}_{i=1}^G = \mathcal{RM}(\{o_i\}_{i=1}^G \,|\, \mathcal{I})
\end{equation}
Here, $\mathcal{RM}$ denotes a reward model that evaluates the entire response group jointly, assigning context-aware and relatively calibrated scores according to the criterion $\mathcal{I}$. 

To address reward hacking due to length bias, we incorporate a \textit{soft overlength penalty} \cite{dapo}. This mechanism discourages overly verbose responses by applying a penalty that grows as the response length exceeds a predefined threshold. The penalty function is defined as:
\begin{equation}
\resizebox{\linewidth}{!}{$
\label{eq:length_control}
r_{\text{length}}(o_i) =
\begin{cases}
0, & |o_i| \leq L_{\text{max}} - L_{\text{cache}} \\
\frac{(L_{\text{max}} - L_{\text{cache}}) - |o_i|}{L_{\text{max}}}, &  L_{\text{max}} - L_{\text{cache}} < |o_i| \leq L_{\text{max}} \\
-1, & |o_i| > L_{\text{max}}
\end{cases}
$}
\end{equation}
Here, $L_{\text{max}}$ denotes the maximum allowed response length, and $L_{\text{cache}}$ defines a buffer interval before the penalty reaches its maximum.

The final reward for each response combines the group-comparative reward and the length penalty:
\begin{equation}
\label{eq:final_reward}
r_i^{\text{final}} = \text{clip}(r_i + r_{\text{length}}(o_i), 0, 1)
\end{equation}

CPO retains the PPO-style policy loss, but plugs in the group-wise comparative reward for the advantage calculation. By introducing explicit comparisons into reward estimation, CPO reduces the \textit{ambiguity} and \textit{instability} inherent in sample-wise scoring, yielding more accurate and stable rewards that reflect true response rankings. As such, CPO is especially effective in subjective, open-ended tasks like role-playing dialogue.

\subsection{CharacterArena}
\label{sec:characterarena}

\begin{figure}[h]
    \centering
    \includegraphics[width=\linewidth]{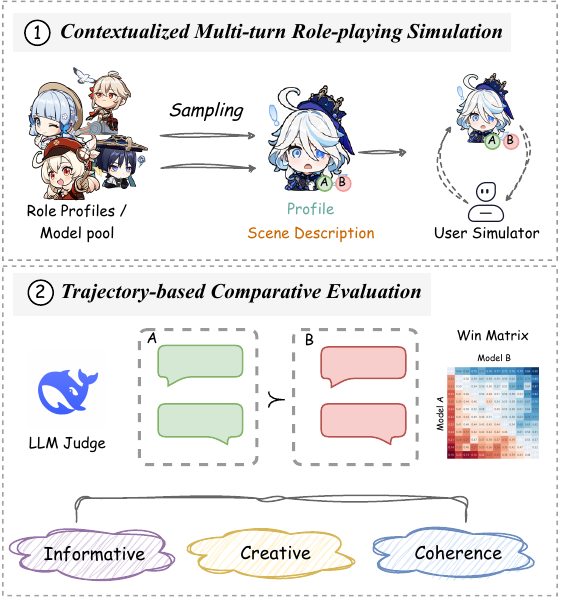}
    \caption{Overview of the CharacterArena framework. It evaluates role-playing agents in two stages: (1) \textit{Contextualized Multi-turn Role-play Simulation} to generate dialogue trajectories, and (2) \textit{Trajectory-based Comparative Evaluation} to reduce context-induced bias.}
    \label{fig:characterarena}
\end{figure}

Existing approaches to evaluating role-playing agents primarily rely on either LLM-based judges or reward models applied to static benchmarks. However, LLM judges often suffer from inconsistent interpretations of evaluation criteria, while reward models typically assess dialogue at the utterance level, overlooking the temporal dependencies inherent in multi-turn conversations. Moreover, both methods are prone to context bias introduced by non-self-generated dialogue history. To overcome these challenges, we present \textit{CharacterArena} - a new evaluation framework that assesses role-playing agents through contextualized dialogue competitions. As illustrated in Figure~\ref{fig:characterarena}, CharacterArena operates in two main phases:

\paragraph{Phase 1: Contextualized Multi-turn Role-play Simulation. }

In this phase, the evaluated models generate complete dialogue trajectories under predefined role-play circumstances. Given a set of models $\mathcal{M} := {m_j}$, we conduct $K$ pairwise matchups for each model pair $(m_A, m_B)$. For each matchup, a chat circumstance $c_i$ is sampled from the set of chat contexts $\mathcal{C}:= {(p_i, s_i)}$, where each $c_i$ comprises a character profile $p_i$ and a scenario $s_i$. Both models then engage in $N$-turn conversations with a shared user simulator $m_{user}$, yielding two dialogue trajectories: $\mathcal{D}_A$ and $\mathcal{D}_B$.

\begin{figure}[t]
    \centering
    \includegraphics[width=\linewidth]{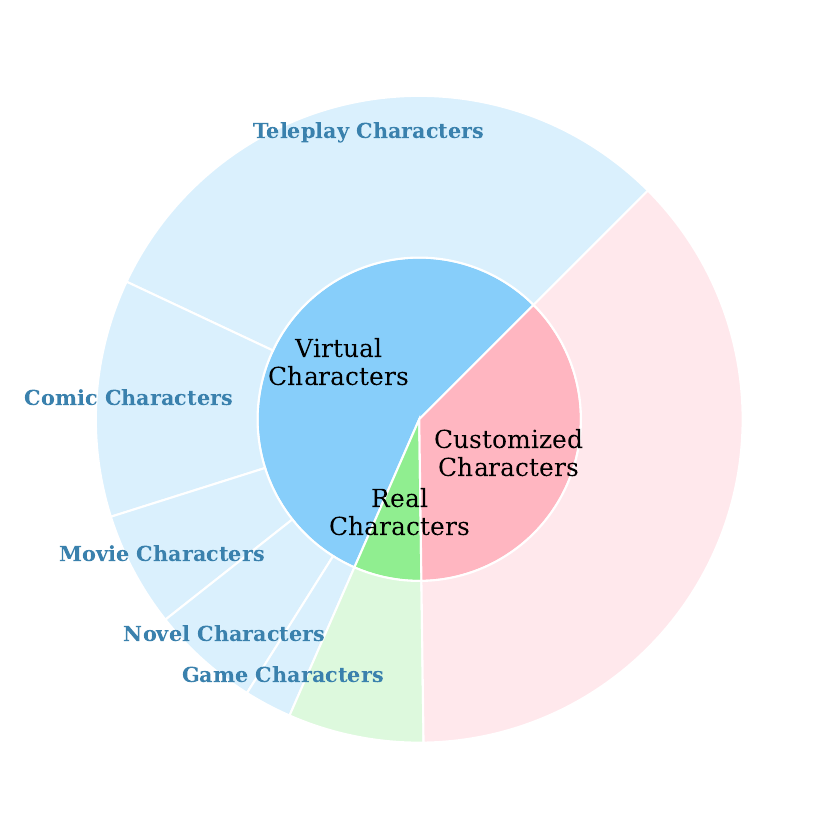}
    \vspace{-5mm}
    \caption{Character distribution. The inner circle shows high-level categories, while the outer circle breaks them down into specific role types.}
    \label{fig:profile_distribution}
\end{figure}

To construct a diverse and meaningful set of chat circumstances, we curate a collection of 294 character profiles spanning virtual personas, historical and public figures, and custom-designed roles. The character distribution is shown in Figure~\ref{fig:profile_distribution}. For each character $p_i$, we employ DeepSeek-R1 \cite{deepseekr1} to generate a scenario $s_i$ that aligns with their background and attributes. An example chat circumstance is provided in Appendix~\ref{apx: Example of Role Profile}.

\paragraph{Phase 2: Trajectory-based Comparative Evaluation.}

In the second phase, we directly compare the generated dialogue trajectories $\mathcal{D}_A$ and $\mathcal{D}_B$ using an LLM judge guided by predefined evaluation criteria. This trajectory-level comparison mitigates biases from local utterance assessment and enables a more holistic evaluation of conversational performance. Results from these pairwise comparisons are aggregated into a win-rate matrix $\mathcal{A}$, which serves as the foundation for model ranking. The evaluation prompt is provided in Appendix \ref{apx:eval_prompts}.

\section{Experimental Setup}
\label{sec:experiment setup}

In this section, we provide a brief overview of the datasets (Section \ref{sec:datasets}), baselines (Section \ref{sec:baseline}), implementation details, and evaluation protocols (Section \ref{sec:evaluation}). The comprehensive experimental setups are available in Appendix \ref{appendix:implementation_details}.

\subsection{Datasets}
\label{sec:datasets} 
\paragraph{SFT Data}
We design two primary tasks in this stage: role-playing and story creation. For the role-playing task, we adopt the RoleplayPref dataset \cite{fft}, which comprises 1,108 unique character roles and 16,888 dialogues. For the story creation task, we use two types of data sources: (1) publicly available novels for story continuation, and (2) the GPT-WritingPrompts dataset\footnote{\href{https://huggingface.co/datasets/vkpriya/GPT-WritingPrompts}{https://huggingface.co/datasets/vkpriya/GPT-WritingPrompts}} for open-ended story generation. In total, the story creation data consists of approximately 50,000 samples.

\paragraph{RLFT Data}
The RLFT stage is primarily aimed at enhancing the model's role-playing capabilities. Consistent with the CharacterArena evaluation protocol, the policy model receives only the character profile during this stage. Training dialogues are generated through interactions with a simulated user model (Doubao-Pro-Character). These profiles are identical to those used in CharacterArena.

\subsection{Baselines}
\label{sec:baseline}
To demonstrate the effectiveness of our approach, we compare it against the
vanilla \textbf{GRPO} \cite{DeepSeekMath}. We employ various LLM backbones to demonstrate the robustness of our method, including \textbf{Qwen2.5 series }(7B, 14B) \cite{qwen2.5} and \textbf{LLaMA3-8B-Instruct} \cite{llama3}. All baselines are trained using the same experimental configuration. Furthermore, we compare our method against the performance of advanced closed-source models, including \textbf{GPT-4o} \cite{gpt-4}, \textbf{Claude-3.7-sonnet} \cite{claude-3-5}, \textbf{Doubao} \cite{doubao}, and \textbf{Minimax} \cite{minimax}.

\subsection{Implementation Details}
\label{implementation}
We sample $N = 16$ responses per context using a \textit{temperature} of $1.0$ and \textit{top-p} of $1.0$ for RLFT. We employ the Qwen2.5-72b-instruct model as the reward model.

\subsection{Evaluation Details}
\label{sec:evaluation}
We evaluate role-playing capabilities from both objective and subjective perspectives. Our objective evaluation comprises three benchmarks: the established utterance-level benchmarks, \textbf{\textit{CharacterEval}} and \textit{\textbf{CharacterBench}}, and our proposed session-level benchmark, \textbf{\textit{CharacterArena}}. To evaluate the subjective performance of the models, we adopt a pairwise human evaluation. Following the setup of CharacterArena, each pair of models engages in multi-turn conversations with a simulated user model under the same dialogue condition. For each pair, 50 dialogue scenarios are simulated, with each scenario consisting of 15 interaction turns. Subsequently, three graduate students independently assessed each dialogue, categorizing the outcome as "A win," "B win," or a "tie". To ensure fairness, we randomize the order of dialogues to eliminate position bias. If all three annotators provide completely different outcomes, the sample is considered invalid and excluded.

\section{Experimental Results}
\label{results}

\subsection{Main results}
\label{main results}

\begin{table*}[th]
\centering
\footnotesize
\setstretch{1.1}
\resizebox{\textwidth}{!}{
\begin{tabular}{lcccccccccccccccc}
\toprule[1.2pt]
\multicolumn{1}{c|}{}                                          & \multicolumn{4}{c|}{\textit{\textbf{Conversational Ability}}}                                                          & \multicolumn{6}{c|}{\textit{\textbf{Character Consistency}}}                                                                                                              & \multicolumn{5}{c|}{\textit{\textbf{Role-playing Attractiveness}}}                                                                              &                                          \\
\multicolumn{1}{c|}{\multirow{-2}{*}{\textit{\textbf{Model}}}} & \textit{\textbf{Flu.}} & \textit{\textbf{Coh.}} & \textit{\textbf{Con.}} & \multicolumn{1}{c|}{\textit{\textbf{Avg.}}} & \textit{\textbf{Exp.}} & \textit{\textbf{Acc.}} & \textit{\textbf{Hall.}} & \textit{\textbf{Beh.}} & \textit{\textbf{Utt.}} & \multicolumn{1}{c|}{\textit{\textbf{Avg.}}} & \textit{\textbf{Hum.}} & \textit{\textbf{Com.}} & \textit{\textbf{Div.}} & \textit{\textbf{Emp.}} & \multicolumn{1}{c|}{\textit{\textbf{Avg.}}} & \multirow{-2}{*}{\textit{\textbf{Avg.}}} \\ \midrule
\multicolumn{17}{l}{\textit{\textbf{Open-source LLMs}}}  \\ \midrule
\rowcolor[HTML]{EFEFEF} 
\multicolumn{1}{l|}{Qwen-2.5-7b-SFT}                           
& 3.51                  
& 3.92              
& 3.71   
&\multicolumn{1}{c|}{3.71}          
& 2.15               
& 2.98      
& 2.97            
& 3.52           
& 3.10            
& \multicolumn{1}{c|}{2.94}          
& 3.57                
& 3.23        
& 2.90               
& 3.12            
& \multicolumn{1}{c|}{3.20}        
& 3.29       \\
\multicolumn{1}{l|}{+ GRPO}                                    
& \cellcolor{lightgreen}3.54                   
& \cellcolor{lightgreen}3.96                   
& \cellcolor{green}3.71        
& \multicolumn{1}{c|}{\cellcolor{lightgreen}3.73}                  
& \cellcolor{lightgreen}2.17            
& \cellcolor{lightgreen}2.99             
& \cellcolor{lightgreen}2.96             
& \cellcolor{lightgreen}3.55               
& \cellcolor{lightgreen}3.10               
& \multicolumn{1}{c|}{\cellcolor{lightgreen}2.95}     
& \cellcolor{green}{3.54}             
& \cellcolor{lightgreen}3.25             
& \cellcolor{lightgreen}2.96             
& \cellcolor{lightgreen}3.13            
& \multicolumn{1}{c|}{\cellcolor{lightgreen}3.22}      
& \cellcolor{lightgreen}3.30   \\
\multicolumn{1}{l|}{+ CPO (Ours)}                              
& {\cellcolor{green}3.58}         
& {\cellcolor{green}3.97}    
& {\cellcolor{green}3.71}    
& \multicolumn{1}{c|}{\cellcolor{green}{3.75}}  
& \cellcolor{green}{2.29}           
& \cellcolor{green}{3.01}            
& {\cellcolor{green}3.03}   
& \cellcolor{green}{3.67}            
& {\cellcolor{green}3.13}     
& \multicolumn{1}{c|}{\cellcolor{green}{3.03}}    
& \cellcolor{lightgreen}3.49     
& \cellcolor{green}3.40              
& \cellcolor{green}{3.07}           
& {\cellcolor{green}3.23}    
& \multicolumn{1}{c|}{\cellcolor{green}{3.30}}           
& {\cellcolor{green}3.36}        \\ 
\hline
\rowcolor[HTML]{EFEFEF} 
\multicolumn{1}{l|}{Qwen-2.5-14b-SFT}                       
& 3.57                
& \textbf{4.04}      
& {3.82}               
& \multicolumn{1}{c|}{{3.81}}         
& 2.22                
& 3.05          
& 3.02                
& 3.67             
& 3.16             
& \multicolumn{1}{c|}{3.02}        
& {3.61}              
& 3.36               
& {3.04}           
& {3.24}             
& \multicolumn{1}{c|}{{3.31}}     
& {3.38}        \\
\multicolumn{1}{l|}{+ GRPO}                  
& \cellcolor{lightgreen}3.59                       
& \cellcolor{green}4.01                      
& \cellcolor{lightgreen}3.79                       
& \multicolumn{1}{c|}{\cellcolor{lightgreen}3.80}                      
&  \cellcolor{lightgreen}2.24                     
&  \cellcolor{lightgreen}3.04              
&  \cellcolor{lightgreen}3.01                   
& \cellcolor{lightgreen}3.65             
& \cellcolor{lightgreen}3.16                
& \multicolumn{1}{c|}{\cellcolor{lightgreen}3.02}      
&  \cellcolor{green}3.60                    
& \cellcolor{lightgreen}3.37                  
& \cellcolor{green}3.03               
& \cellcolor{green}3.22                
& \multicolumn{1}{c|}{\cellcolor{green}{3.31}}       
&  \cellcolor{lightgreen}3.37    \\
\multicolumn{1}{l|}{+ CPO (Ours)}            
& \cellcolor{green}{3.62}               
&  \cellcolor{lightgreen}4.00               
&  \cellcolor{green}3.80               
& \multicolumn{1}{c|}{\cellcolor{green}{3.81}}   
& \cellcolor{green}{2.25}                 
& \cellcolor{green}{3.06}                
& \cellcolor{green}{3.05}                
& \cellcolor{green}{3.69}               
& \cellcolor{green}{3.17}             
& \multicolumn{1}{c|}{\cellcolor{green}{3.04}}       
& \cellcolor{green}3.60               
& \cellcolor{green}{3.39}             
& \cellcolor{lightgreen}2.99                  
&  \cellcolor{lightgreen}3.20               
& \multicolumn{1}{c|}{\cellcolor{lightgreen}3.29}       
& \cellcolor{green}{3.38}      \\ 
\hline
\rowcolor[HTML]{EFEFEF} 
\multicolumn{1}{l|}{LLaMA-3-8b-SFT}                            & 3.38                   & 3.80                   & 3.48                   & \multicolumn{1}{c|}{3.55}                   & 2.19                   & 2.90                   & 2.85                    & 3.64                   & 3.00                   & \multicolumn{1}{c|}{2.92}                   & 3.24                   & 3.21                   & 3.06                   & 3.02                   & \multicolumn{1}{c|}{3.13}                   & 3.20                                     \\
\multicolumn{1}{l|}{+ GRPO}                
& \cellcolor{lightgreen}3.35            
& \cellcolor{lightgreen}3.79           
& \cellcolor{green}3.48                 
& \multicolumn{1}{c|}{\cellcolor{lightgreen}3.54}          
& \cellcolor{lightgreen}2.10              
& \cellcolor{lightgreen}2.87                    
&  \cellcolor{lightgreen}2.82                  
& \cellcolor{lightgreen}3.52                   
&  \cellcolor{lightgreen}2.98                  
& \multicolumn{1}{c|}{\cellcolor{lightgreen}2.86}     
& \cellcolor{green}3.32                 
& \cellcolor{lightgreen}3.08               
&  \cellcolor{lightgreen}2.92              
&  \cellcolor{lightgreen}2.97              
& \multicolumn{1}{c|}{\cellcolor{lightgreen}3.07}            
&   \cellcolor{lightgreen}3.16         \\
\multicolumn{1}{l|}{+ CPO (Ours)}     
& \cellcolor{green}3.36          
& \cellcolor{green}3.81        
& \cellcolor{green}3.48           
& \multicolumn{1}{c|}{\cellcolor{green}3.55}     
& \cellcolor{green}2.19               
& \cellcolor{green}2.90             
&\cellcolor{green}2.87                
& \cellcolor{green}3.62               
& \cellcolor{green}2.99           
& \multicolumn{1}{c|}{\cellcolor{green}2.91}       
& \cellcolor{lightgreen}3.26   
& \cellcolor{green}3.20              
& \cellcolor{green}3.02             
& \cellcolor{green}3.04             
& \multicolumn{1}{c|}{\cellcolor{green}3.13}      
& \cellcolor{green}3.20\\ 
\midrule \midrule
\multicolumn{17}{l}{\textit{\textbf{Close-source LLMs}}}   \\ \midrule
\multicolumn{1}{l|}{MiniMax-abab5.5s}                          & 3.61                   & 3.93                   & 3.81                   & \multicolumn{1}{c|}{3.78}                   & 1.84                   & 2.91                   & 2.94                    & 2.77                   & 3.13                   & \multicolumn{1}{c|}{2.72}                   & 3.77                   & 2.67                   & 2.15                   & 3.01                   & \multicolumn{1}{c|}{2.90}                   & 3.13                                     \\
\multicolumn{1}{l|}{Deepseek-R1}                               & 3.53                   & 3.83                   & 3.96                   & \multicolumn{1}{c|}{3.77}                   & 1.34                   & 3.07                   & 2.88                    & 1.70                   & 3.21                   & \multicolumn{1}{c|}{2.44}                   & \textbf{4.38}          & 1.71                   & 1.56                   & 2.93                   & \multicolumn{1}{c|}{2.64}                   & 2.95                                     \\
\multicolumn{1}{l|}{GPT-4o}                                    & 3.54                   & 3.89                   & 3.47                   & \multicolumn{1}{c|}{3.63}                   & 2.58                   & 3.13                   & 2.99                    & 2.83                   & 2.98                   & \multicolumn{1}{c|}{2.90}                   & 3.17                   & 3.54                   & 2.20                   & 3.32                   & \multicolumn{1}{c|}{3.06}                   & 3.20                                     \\
\multicolumn{1}{l|}{Doubao-Pro-Character}                      & 3.61                   & 3.94                   & 3.64                   & \multicolumn{1}{c|}{3.73}                   & \textbf{2.85}          & \textbf{3.41}          & \textbf{3.17}           & 3.87                   & 3.17                   & \multicolumn{1}{c|}{\textbf{3.29}}          & 3.42                   & \textbf{3.74}          & 3.35                   & \textbf{3.53}          & \multicolumn{1}{c|}{\textbf{3.51}}          & \textbf{3.51}                            \\
\multicolumn{1}{l|}{Claude-3.7-sonnet}                         & \textbf{3.71}          & {3.99}          & \textbf{4.00}          & \multicolumn{1}{c|}{\textbf{3.90}}          & 2.03                   & 3.06                   & 3.04                    & \textbf{3.91}          & \textbf{3.26}          & \multicolumn{1}{c|}{3.06}                   & 3.91                   & 2.75                   & \textbf{3.38}          & 2.97                   & \multicolumn{1}{c|}{3.25}                   & 3.41                                     \\ \bottomrule[1.2pt]
\end{tabular}
}
\caption{
\textbf{The CharacterEval Benchmark} \cite{tu-etal-2024-charactereval}. 
The best and second-best scores in different rewarding paradigms are highlighted in \colorbox{green}{``Green''} and \colorbox{lightgreen}{``Lightgreen''}.  
The best result is shown in \textbf{bold}.
}
\label{tab:character_eval}
\end{table*}

\begin{table*}[th]
\centering
\footnotesize
\setstretch{1.1}
\resizebox{\textwidth}{!}{%
\begin{tabular}{lcccccccccccccc}
\toprule[1.2pt]
    \multicolumn{15}{l}{\makecell[l]{
    \textbb{MC}: Memory Consistency ~~~ 
    \textbb{FA}: Fact Accuracy ~~~ 
    \textbb{BC$_K$}: Boundary Consistency ~~~ 
    \textbb{AC$^h$}: Attribute Consistency (Human) ~~~
    \textbb{EG}: Engagement \\
    \textbb{AC$^b$}: Attribute Consistency (Bot)~~~
    \textbb{BC$_P^b$}: Behavior Consistency (Bot) ~~~ 
    \textbb{BC$_P^h$}: Behavior Consistency (Human) ~~~
    \textbb{HL}: Human-likeness 
    \\
    \textbb{ES}: Emotional Self-regulation ~~~
    \textbb{ER}: Empathetic Responsiveness ~~~
    \textbb{MS}: Morality Stability ~~~ 
    \textbb{MR}: Morality Robustness
    }} \\
\hline
\multicolumn{1}{c|}{}                                         & \multicolumn{1}{c|}{}                                   & \multicolumn{1}{c|}{\textbf{\textit{Memory}}}              & \multicolumn{2}{c|}{\textbf{\textit{Knowledge}}}                                  & \multicolumn{4}{c|}{\textbf{\textit{Person}}}                                                                                                 & \multicolumn{2}{c|}{\textbf{\textit{Emotion}}}                                    & \multicolumn{2}{c|}{\textbf{\textit{Morality}}}                                   & \multicolumn{2}{c}{\textbf{\textit{Believability}}}  \\
\multicolumn{1}{c|}{\multirow{-2}{*}{\textbf{\textit{Model}}}}         & \multicolumn{1}{c|}{\multirow{-2}{*}{\textbf{\textit{Average}}}} & \multicolumn{1}{c|}{\textbf{\textbb{MC}}}         & \textbf{\textbb{FA}} & \multicolumn{1}{c|}{\textbf{\textbb{BC$_K$}}}     & \textbf{\textbb{AC$^b$}} & \textbf{\textbb{AC$^h$}} & \textbf{\textbb{BC$_P^b$}} & \multicolumn{1}{c|}{\textbf{\textbb{BC$_P^h$}}}   & \textbf{\textbb{ES}} & \multicolumn{1}{c|}{\textbf{\textbb{ER}}}         & \textbf{\textbb{MS}} & \multicolumn{1}{c|}{\textbf{\textbb{MR}}}         & \textbf{\textbb{HL}} & \textbf{\textbb{EG}} \\ \midrule
\multicolumn{15}{l}{\textit{\textbf{Open-source LLMs}}}  \\ \midrule
\rowcolor[HTML]{F3F3F3} 
\multicolumn{1}{l|}{\cellcolor[HTML]{F3F3F3}Qwen-2.5-7b-SFT}  & \multicolumn{1}{c|}{\cellcolor[HTML]{F3F3F3}3.33}       & \multicolumn{1}{c|}{\cellcolor[HTML]{F3F3F3}3.15} & 2.32                 & \multicolumn{1}{c|}{\cellcolor[HTML]{F3F3F3}3.59} & 3.33                     & 3.51                     & 3.17                       & \multicolumn{1}{c|}{\cellcolor[HTML]{F3F3F3}3.12} &3.04                 & \multicolumn{1}{c|}{\cellcolor[HTML]{F3F3F3}2.68} & 4.68                 & \multicolumn{1}{c|}{\cellcolor[HTML]{F3F3F3}4.76} & 2.84                 & 3.13                 \\
\multicolumn{1}{l|}{+ GRPO}                               
& \multicolumn{1}{c|}{\cellcolor{lightgreen}3.41}             
& \multicolumn{1}{c|}{\cellcolor{lightgreen}3.29}          
& \cellcolor{lightgreen}2.42            
& \multicolumn{1}{c|}{\cellcolor{lightgreen}3.74}        
& \cellcolor{lightgreen}3.56                
& \cellcolor{lightgreen}3.69  
& \cellcolor{lightgreen}3.14                 
& \multicolumn{1}{c|}{\cellcolor{green}3.14}     
& \cellcolor{lightgreen}3.03           
& \multicolumn{1}{c|}{\cellcolor{lightgreen}2.73}      
& \cellcolor{lightgreen}4.76              
& \multicolumn{1}{c|}{\cellcolor{green}4.80}     
& \cellcolor{green}2.94           
&\cellcolor{green}3.12                 \\
\multicolumn{1}{l|}{+ CPO (Ours)}                     
& \multicolumn{1}{c|}{\cellcolor{green}3.49}           
& \multicolumn{1}{c|}{\cellcolor{green}3.44}          
& \cellcolor{green}2.48         
& \multicolumn{1}{c|}{\cellcolor{green}3.78}      
& \cellcolor{green}3.75             
& \cellcolor{green}3.91            
& \cellcolor{green}3.38            
& \multicolumn{1}{c|}{\cellcolor{lightgreen}3.11}       
& \cellcolor{green}3.09            
& \multicolumn{1}{c|}{\cellcolor{green}2.81}   
& \cellcolor{green}4.91           
& \multicolumn{1}{c|}{\cellcolor{lightgreen}4.77}     
& \cellcolor{lightgreen}2.84           
& \cellcolor{lightgreen}3.11                 \\ \hline
\rowcolor[HTML]{F3F3F3} 
\multicolumn{1}{l|}{\cellcolor[HTML]{F3F3F3}Qwen-2.5-14b-SFT} & \multicolumn{1}{c|}{\cellcolor[HTML]{F3F3F3}3.57}       & \multicolumn{1}{c|}{\cellcolor[HTML]{F3F3F3}3.53} & 2.59                 & \multicolumn{1}{c|}{\cellcolor[HTML]{F3F3F3}3.85} & 3.81                     & 3.90                     & 3.38                       & \multicolumn{1}{c|}{\cellcolor[HTML]{F3F3F3}3.22} & 3.17                 & \multicolumn{1}{c|}{\cellcolor[HTML]{F3F3F3}2.93} & 4.83                 & \multicolumn{1}{c|}{\cellcolor[HTML]{F3F3F3}4.84} & 3.12                 & 3.30                 \\
\multicolumn{1}{l|}{+ GRPO}                                
& \multicolumn{1}{c|}{\cellcolor{lightgreen}3.55}               
& \multicolumn{1}{c|}{\cellcolor{lightgreen}3.38}      
& \cellcolor{lightgreen}2.48                 
& \multicolumn{1}{c|}{\cellcolor{green}3.77}          
&\cellcolor{green}3.69                 
& \cellcolor{green}3.94                 
& \cellcolor{lightgreen}3.35                   
& \multicolumn{1}{c|}{\cellcolor{lightgreen}3.26}           
& \cellcolor{lightgreen}3.17                
& \multicolumn{1}{c|}{\cellcolor{lightgreen}2.94}           
& \cellcolor{lightgreen}4.84               
& \multicolumn{1}{c|}{\cellcolor{lightgreen}4.76}           
& \cellcolor{green}3.24               
& \cellcolor{lightgreen}3.37                     \\
\multicolumn{1}{l|}{+ CPO (Ours)}                          
& \multicolumn{1}{c|}{\cellcolor{green}3.59}            
& \multicolumn{1}{c|}{\cellcolor{green}3.48}     
&\cellcolor{green}2.55               
&\multicolumn{1}{c|}{\cellcolor{green}3.77}       
&\cellcolor{green}3.69                    
&\cellcolor{lightgreen}3.92                    
& \cellcolor{green}3.42                    
& \multicolumn{1}{c|}{\cellcolor{green}3.30}    
& \cellcolor{green}3.36                
& \multicolumn{1}{c|}{\cellcolor{green}2.99}      
& \cellcolor{green}4.90               
& \multicolumn{1}{c|}{\cellcolor{green}4.85}         
& \cellcolor{lightgreen}3.03                
& \cellcolor{green}3.39                      \\ \hline
\rowcolor[HTML]{F3F3F3} 
\multicolumn{1}{l|}{\cellcolor[HTML]{F3F3F3}LLaMA-3-8b-SFT}   & \multicolumn{1}{c|}{\cellcolor[HTML]{F3F3F3}3.29}       & \multicolumn{1}{c|}{\cellcolor[HTML]{F3F3F3}3.10} & 2.24                 & \multicolumn{1}{c|}{\cellcolor[HTML]{F3F3F3}3.58} & 3.43                     & 3.58                     & 3.17                       & \multicolumn{1}{c|}{\cellcolor[HTML]{F3F3F3}3.15} & 2.86                 & \multicolumn{1}{c|}{\cellcolor[HTML]{F3F3F3}2.70} & 4.64                 & \multicolumn{1}{c|}{\cellcolor[HTML]{F3F3F3}4.63} & 2.73                 & 2.92                 \\
\multicolumn{1}{l|}{+ GRPO}                                   
& \multicolumn{1}{c|}{\cellcolor{lightgreen}3.31}                                   
& \multicolumn{1}{c|}{\cellcolor{green}3.19}                             
& \cellcolor{green}2.23                     
& \multicolumn{1}{c|}{\cellcolor{lightgreen}3.68}                             
& \cellcolor{green}3.60                         
&  \cellcolor{lightgreen}3.55                        
& \cellcolor{green}3.23                          
& \multicolumn{1}{c|}{\cellcolor{lightgreen}3.04}                           
& \cellcolor{lightgreen}2.85                   
& \multicolumn{1}{c|}{\cellcolor{lightgreen}2.66}           
& \cellcolor{green}4.67                    
& \multicolumn{1}{c|}{\cellcolor{lightgreen}4.65}                          
& \cellcolor{lightgreen}2.73           
&\cellcolor{lightgreen}2.98                      \\
\multicolumn{1}{l|}{+ CPO (Ours)}                             
& \multicolumn{1}{c|}{\cellcolor{green}3.36}                                   
& \multicolumn{1}{c|}{\cellcolor{lightgreen}3.17}                             
& \cellcolor{green}2.23                    
& \multicolumn{1}{c|}{\cellcolor{green}3.82}                             
&  \cellcolor{lightgreen}3.54                        
&  \cellcolor{green}3.63                       
&  \cellcolor{lightgreen}3.21                          
& \multicolumn{1}{c|}{\cellcolor{green}3.21}                             
& \cellcolor{green}2.90                     
& \multicolumn{1}{c|}{\cellcolor{green}2.68}                             
&  \cellcolor{lightgreen}4.60                    
& \multicolumn{1}{c|}{\cellcolor{green}4.76}                             
& \cellcolor{green}2.92                    
& \cellcolor{green}3.02                    \\ \midrule \midrule
\multicolumn{15}{l}{\textit{\textbf{Closed-source LLMs}}} \\ \midrule
\multicolumn{1}{l|}{MiniMax-abab5.5s}                         & \multicolumn{1}{c|}{3.52}                               & \multicolumn{1}{c|}{3.76}                         & 2.76                 & \multicolumn{1}{c|}{3.45}                         & 4.18                     & 4.02                     & 3.35                       & \multicolumn{1}{c|}{3.04}                         & 3.04                 & \multicolumn{1}{c|}{2.71}                         & 4.69                 & \multicolumn{1}{c|}{4.65}                         & 3.02                 & 3.15                 \\
\multicolumn{1}{l|}{Deepseek-R1}                              & \multicolumn{1}{c|}{3.77}                               & \multicolumn{1}{c|}{3.56}                         & 3.17                 & \multicolumn{1}{c|}{3.76}                         & 4.28                     & 4.21                     & 3.81                       & \multicolumn{1}{c|}{4.00}                         & 3.17                 & \multicolumn{1}{c|}{3.08}                         & 4.69                 & \multicolumn{1}{c|}{4.53}                         & 3.20                 & 3.49                 \\
\multicolumn{1}{l|}{GPT-4o}                                   & \multicolumn{1}{c|}{3.86}                               & \multicolumn{1}{c|}{3.83}                         & \textbf{3.28}        & \multicolumn{1}{c|}{3.86}                         & \textbf{4.73}            & 4.38                     & 3.81                       & \multicolumn{1}{c|}{3.60}                         & 3.53                 & \multicolumn{1}{c|}{3.51}                         & \textbf{4.96}        & \multicolumn{1}{c|}{\textbf{4.91}}                & 2.68                 & 3.06                 \\
\multicolumn{1}{l|}{Doubao-Pro-Character}                     & \multicolumn{1}{c|}{3.90}                               & \multicolumn{1}{c|}{3.67}                         & 3.04                 & \multicolumn{1}{c|}{3.98}                         & 4.49                     & 4.53                     & 3.88                       & \multicolumn{1}{c|}{3.92}                         & 3.31                 & \multicolumn{1}{c|}{3.17}                         & 4.91                 & \multicolumn{1}{c|}{\textbf{4.91}}                & \textbf{3.40}        & 3.49                 \\
\multicolumn{1}{l|}{Claude-3.7-sonnet}                        & \multicolumn{1}{c|}{\textbf{4.07}}                      & \multicolumn{1}{c|}{\textbf{4.03}}                & 3.12                 & \multicolumn{1}{c|}{\textbf{4.21}}                & \textbf{4.73}            & \textbf{4.60}            & \textbf{4.03}              & \multicolumn{1}{c|}{\textbf{4.34}}                & \textbf{3.84}        & \multicolumn{1}{c|}{\textbf{3.61}}                & 4.94                 & \multicolumn{1}{c|}{4.83}                         & 3.14                 & \textbf{3.52}        \\ \bottomrule[1.2pt]
\end{tabular}
}
\caption{
\textbf{The CharacterBench (zh) Benchmark} \cite{characterbench}. 
}
\label{tab:character_bench_zh}
\end{table*}

\paragraph{Utterance-level Benchmark Results}
Tables ~\ref{tab:character_eval} and ~\ref{tab:character_bench_zh} present the results on \textbf{CharacterEval} and \textbf{CharacterBench}, respectively. Our analysis reveals several key findings: 
(1) {\textbf{CPO consistently outperforms GRPO across multiple evaluation dimensions}}. On CharacterEval, CPO achieves superior scores in Conversational Ability, Character Consistency, and Role-playing Attractiveness. Similar gains are observed on CharacterBench, where CPO shows consistent improvements across most metrics.
(2) {\textbf{CPO delivers superior and stable performance across diverse backbone architectures}}. On CharacterEval, CPO surpasses GRPO by an average of 0.06 on Qwen-2.5-7b and 0.04 on LLaMA-3-8b. On CharacterBench, it achieves average gains of 0.08 on Qwen-2.5-7b, 0.04 on Qwen-2.5-14b, and 0.05 on LLaMA-3-8b. These results highlight CPO’s strong generalization capabilities and architecture-agnostic effectiveness.

\paragraph{Session-level Benchmark Results}
\begin{figure}[th]
    \centering
    \begin{subfigure}{0.45\linewidth}
        \centering
        \includegraphics[width=\textwidth]{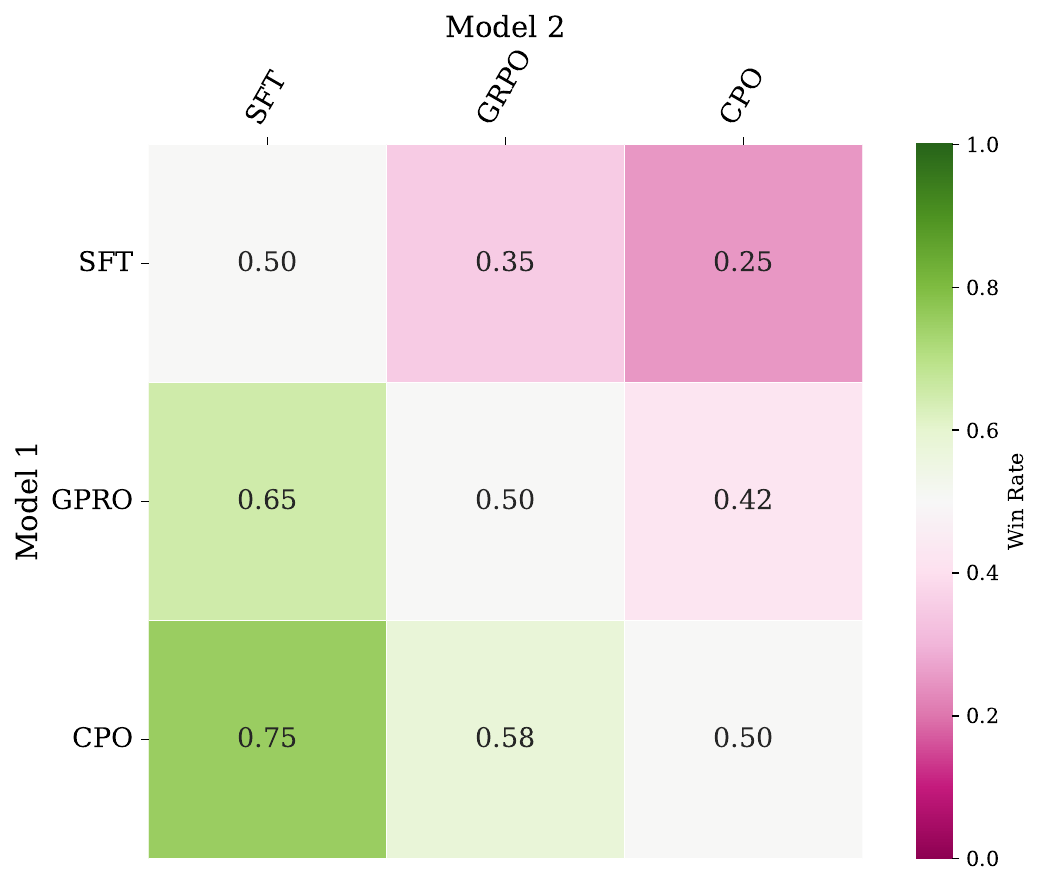}
        \vspace{-3mm}
        \caption{Qwen-2.5-7b-Instruct}
    \end{subfigure}
    \begin{subfigure}{0.45\linewidth}
        \centering
        \includegraphics[width=\textwidth]{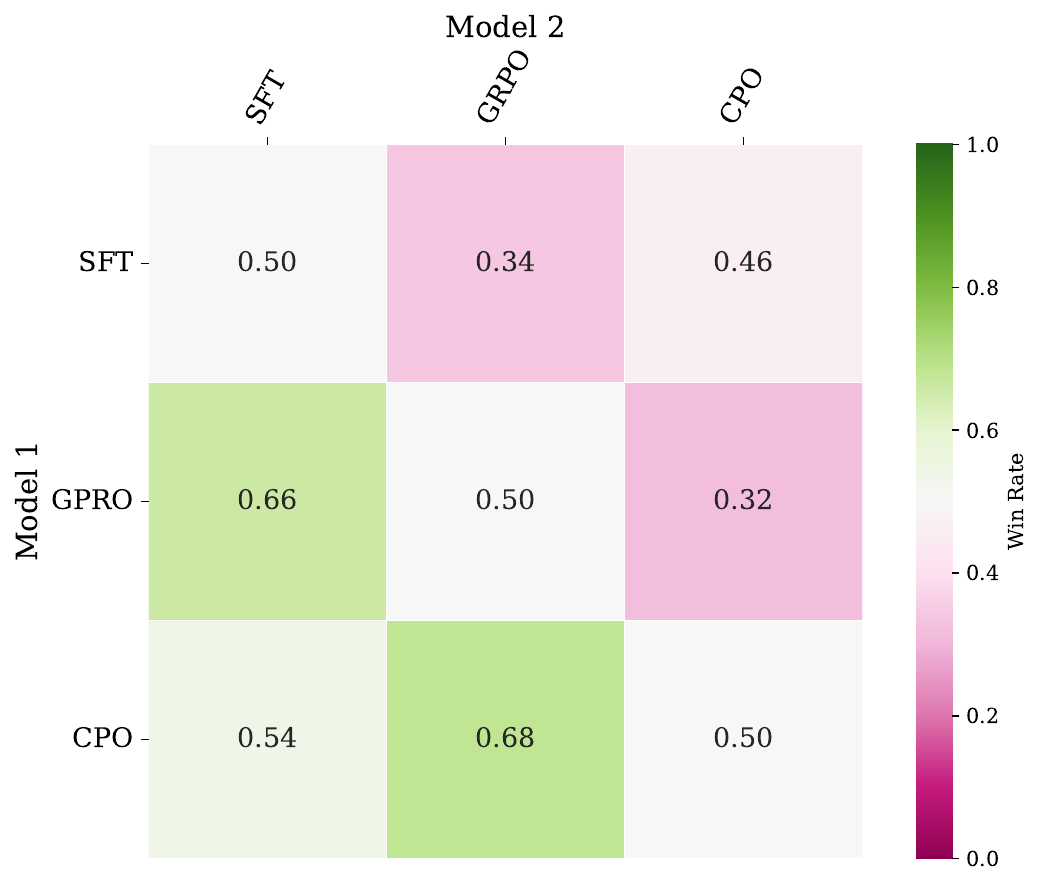}
        \vspace{-3mm}
        \caption{Qwen-2.5-14b-Instruct}
    \end{subfigure}
    \begin{subfigure}{0.45\linewidth}
        \centering
        \includegraphics[width=\textwidth]{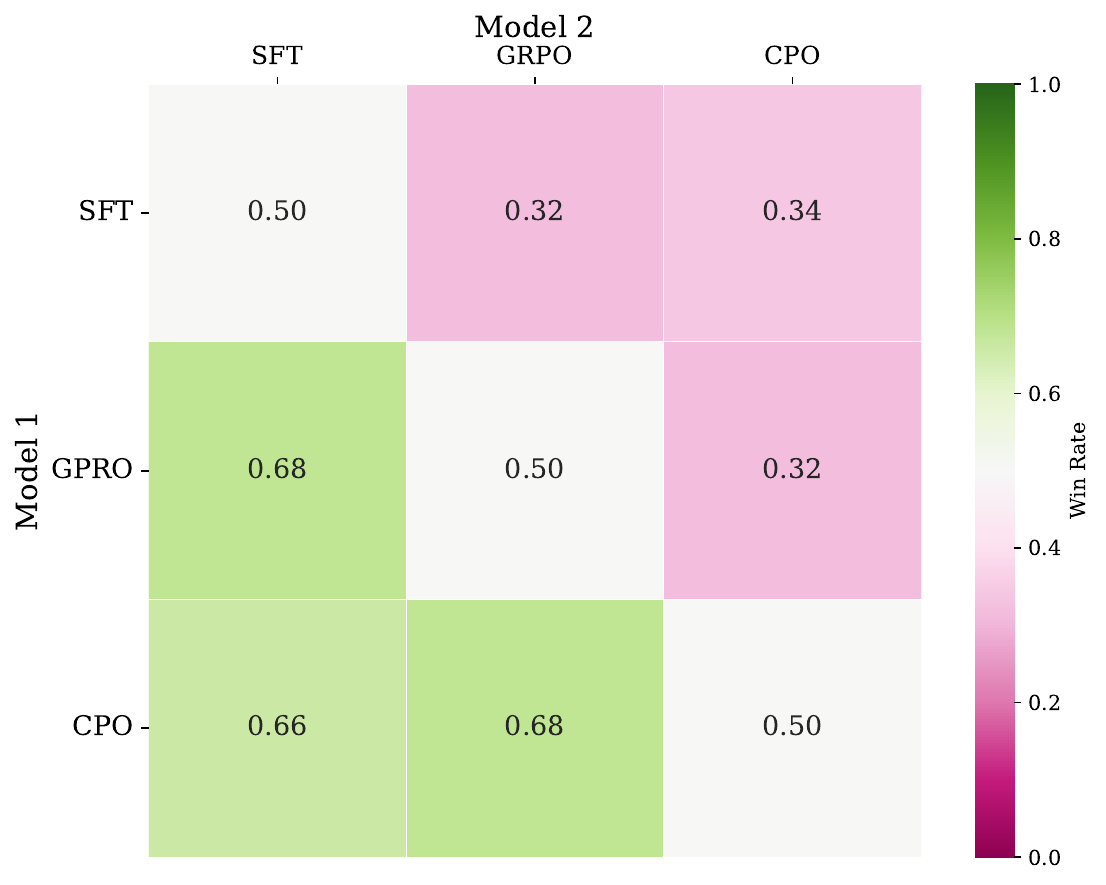}
        \caption{LLaMA-3-8b-Instruct}
    \end{subfigure}
    \begin{subfigure}{0.45\linewidth}
        \centering
        \includegraphics[width=\textwidth]{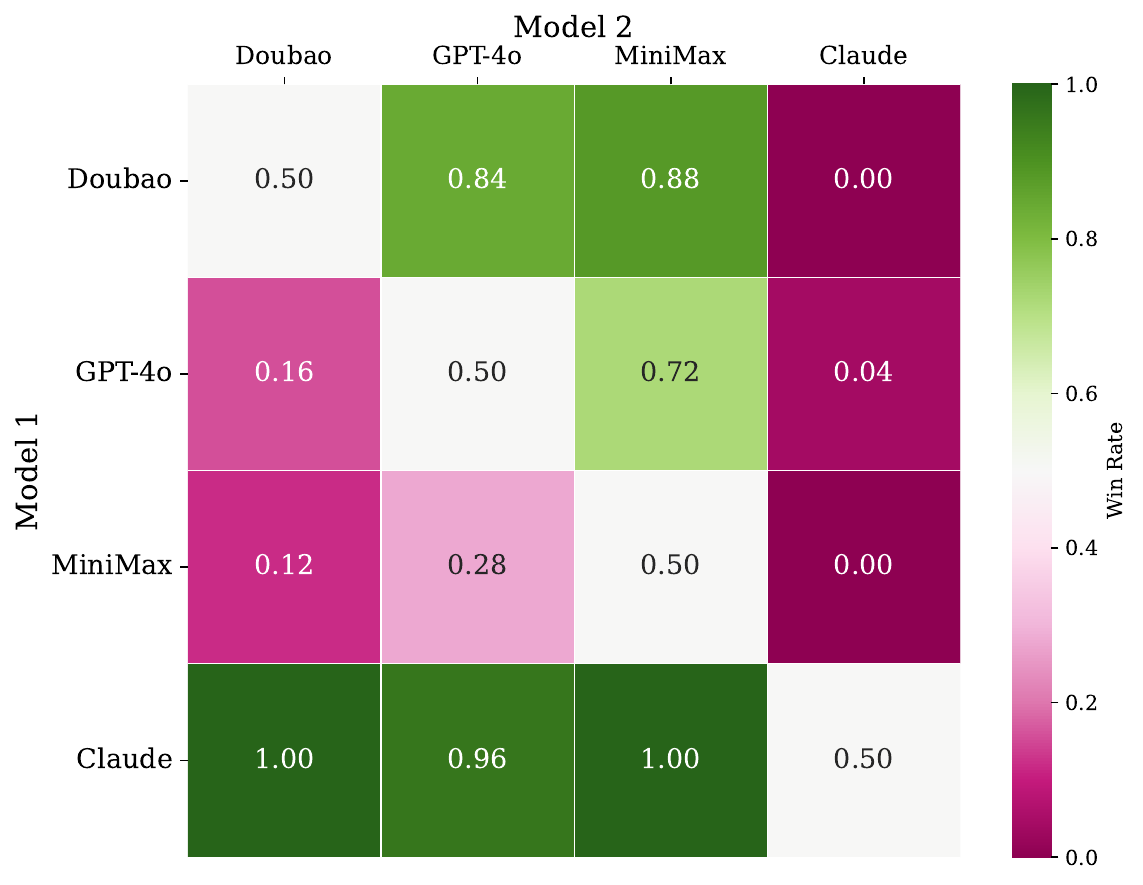}
        \caption{Close Source LLMs}
    \end{subfigure}
    \caption{Win Rate Matrices on the CharacterArena Benchmark. Values in cell $(i,j)$ indicate the preference rate of Model $i$ over Model $j$.}
    \label{fig:character_arena}
\end{figure}

Figure~\ref{fig:character_arena} presents the win rate matrices from the CharacterArena evaluation. The results demonstrate that:
(1) \textbf{{CPO consistently outperforms both SFT and GRPO.}} On all backbones—including Qwen-2.5-7b, Qwen-2.5-14b, and LLaMA-3-8b—CPO achieves higher win rates, underscoring its effectiveness in enhancing role-playing quality.
(2) \textbf{Results from CharacterArena align closely with CharacterEval and CharacterBench}, reinforcing the conclusion that CPO substantially improves the quality and appeal of role-playing agents.

\paragraph{Human Evaluation Results}
\begin{figure}[th]
    \centering
    \includegraphics[width=0.8\linewidth]{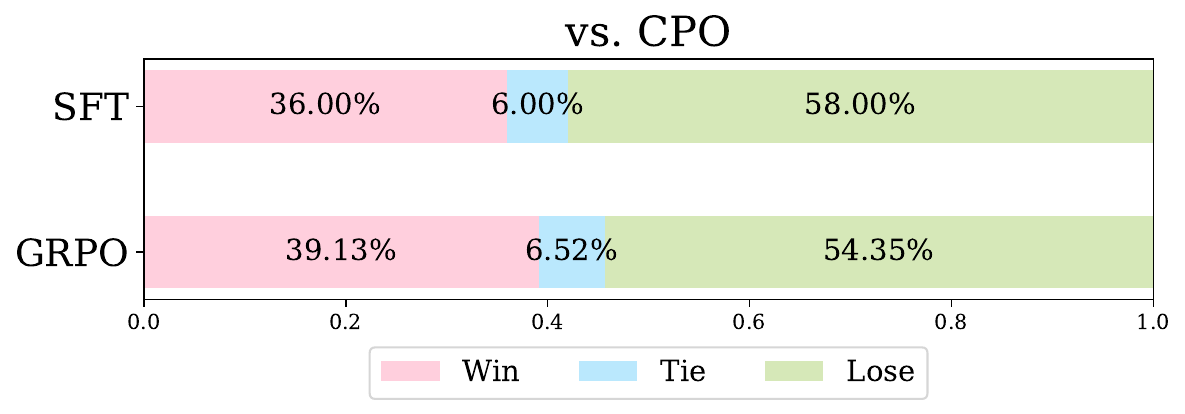}
    \caption{Pairwise human evaluation results on \textit{Qwen-2.5-7b-instruct}. In the `A vs B' comparisons, {\color[HTML]{ffb0c7}$\blacksquare$} indicates `A win', {\color[HTML]{54c6fc}$\blacksquare$} indicates `tie', and {\color[HTML]{bcd989}$\blacksquare$} indicates `B win'. \textit{The Fleiss' Kappa score is 0.473}.}
    \label{fig:human_evaluation}
    \vspace{-6mm}
\end{figure}

As shown in Figure~\ref{fig:human_evaluation}, CPO achieves higher win rates than both SFT and GRPO. However, compared to its performance on CharacterArena, the win rate of CPO in human evaluation appears more conservative. 
This discrepancy can be partly attributed to the moderate inter-annotator agreement (Fleiss’ Kappa = 0.473), suggesting considerable variance in human preferences and subjective judgment criteria. These findings further underscore the inherent ambiguity and difficulty of role-playing evaluation.

\subsection{Analysis and Discussion}

This section investigates the following questions:

\noindent  \textbf{Q1:} Is group-wise rewarding more effective than sample-wise rewarding?

\noindent  \textbf{Q2:} Does group-wise rewarding generalize well to other RLFT methods?

\noindent  \textbf{Q3:} Is LLM-based evaluation in CharacterArena reliable?

\subsubsection{Efficacy of Group-wise Rewarding}

\begin{figure}[th]
    \centering
    \includegraphics[width=0.9\linewidth]{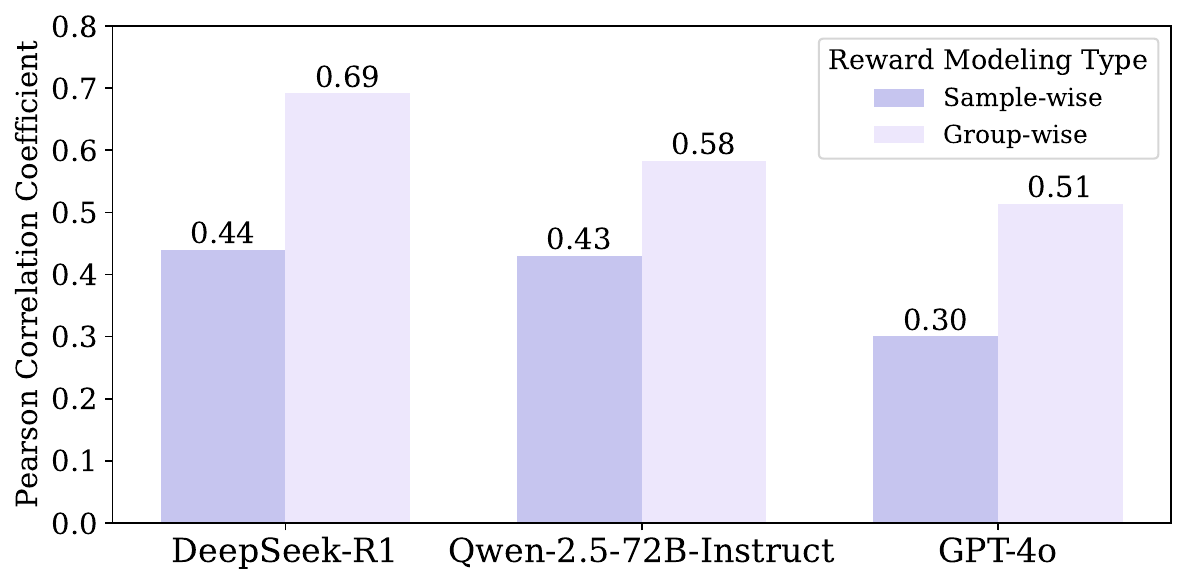}
    \caption{Pearson Correlation Coefficients between human and LLM judges for different reward scoring modes, comparing sample-wise and group-wise approaches.}
    \label{fig:pearson_of_human}
\end{figure}

To evaluate the effectiveness of our group-wise rewarding strategy compared to the traditional sample-wise rewarding, we compute the Pearson correlation coefficients between LLM-generated scores and human annotations for both approaches. Specifically, we employ three LLM judge models: DeepSeek-R1, Qwen-2.5-72b-Instruct, and GPT-4o. We construct the evaluation set by selecting 50 dialogue contexts. For each context, five candidate responses are generated by the policy model based on the dialogue history. These responses are then scored by both humans and LLM judges.

As shown in Figure~\ref{fig:pearson_of_human}, the group-wise rewarding method consistently achieves higher correlation with human judgments across all three models. Specifically, group-wise scoring yields a 25\% improvement in correlation over sample-wise scoring on DeepSeek-R1, 21\% on GPT-4o, and 15\% on Qwen-2.5-72b. These results demonstrate that group-wise rewarding more effectively aligns LLM evaluation with human preferences.

\subsubsection{Generalizability of Group-wise Rewarding to Other RLFT Methods}

\begin{table}[t]
\small
\setstretch{1.2}
\centering
\resizebox{\linewidth}{!}{%
\begin{tabular}{l|cccc|c}
\toprule[1.2pt]
\multicolumn{1}{c|}{} & \multicolumn{4}{c|}{\textbf{CharacterEval}} &  \\
\multicolumn{1}{c|}{\multirow{-2}{*}{\textbf{Model}}} & \textbf{\textbb{CA}} & \textbf{\textbb{CC}} & \textbf{\textbb{RPA}} & \textbf{Avg.} & \multirow{-2}{*}{\textbf{\begin{tabular}[c]{@{}c@{}}Character\\ Bench\end{tabular}}} \\ \midrule
\rowcolor[HTML]{EFEFEF} 
Qwen-2.5-7b-SFT & 3.71 & 2.94 & 3.20 & 3.29 & 3.33 \\
+ RFT (s) & 3.68 & 2.97 & 3.21 & 3.29 & 3.44 \\
+ RFT (g) & 3.65 & 3.00 & 3.24 & 3.30 & 3.51 \\ \cdashline{1-6}
+ DPO (s) & 3.34 & 3.19 & 3.35 & 3.29 & 3.47 \\
+ DPO (g) & 3.70 & 3.24 & 3.48 & 3.47 & 3.54 \\
\midrule \midrule
\rowcolor[HTML]{EFEFEF} 
LLaMA-3-8B-SFT & 3.55 & 2.92 & 3.13 & 3.20 & 3.29 \\
+ RFT (s) & 3.33 & 2.87 & 3.05 & 3.08 & 3.27 \\
+ RFT (g) & 3.25 & 2.87 & 3.05 & 3.06 & 3.21 \\
\cdashline{1-6}
+ DPO (s) & 3.63 & 3.20 & 3.48 & 3.44 & 3.54 \\
+ DPO (g) & 3.63 & 3.29 & 3.54 & 3.49 & 3.54 \\
\bottomrule[1.2pt]
\end{tabular}%
}
\caption{Comparison of group-wise and sample-wise rewarding applied to other RLFT methods. `\textit{g}' and `\textit{s}' denote group-wise and sample-wise rewarding, respectively. Metrics reported include \textbb{CA} (Conversational Ability), \textbb{CC} (Character Consistency), and \textbb{RPA} (Role-playing Attractiveness) on CharacterEval and the average results on CharacterBench.}
\label{tab:additional}
\end{table}

Section~\ref{main results} establishes CPO's superiority over GRPO, attributed primarily to its group-wise rewarding strategy. To evaluate its broader applicability, we extend this approach to two RLFT paradigms: RFT and DPO. For RFT, we select the highest-reward response from each group as the training target, while DPO forms preference pairs between the highest- and lowest-scoring responses within groups. 

Experimental results in Table~\ref{tab:additional} demonstrate consistent performance gains across both methods. Notably, group-wise rewarding RFT(g) and DPO(g) outperform their sample-wise rewarding RFT(s) and DPO(s) on Qwen-2.5-7b. The enhancement is particularly pronounced for DPO, Qwen-2.5-7b DPO(g) achieves average score increases of 0.18 on CharacterEval and 0.07 on CharacterBench (zh) compared to DPO(s). These results confirm the generalizability of group-wise rewarding across RLFT frameworks.

\subsubsection{Reliability of LLM-based Evaluation in CharacterArena}

\begin{figure}[th]
    \centering
    \includegraphics[width=0.9\linewidth]{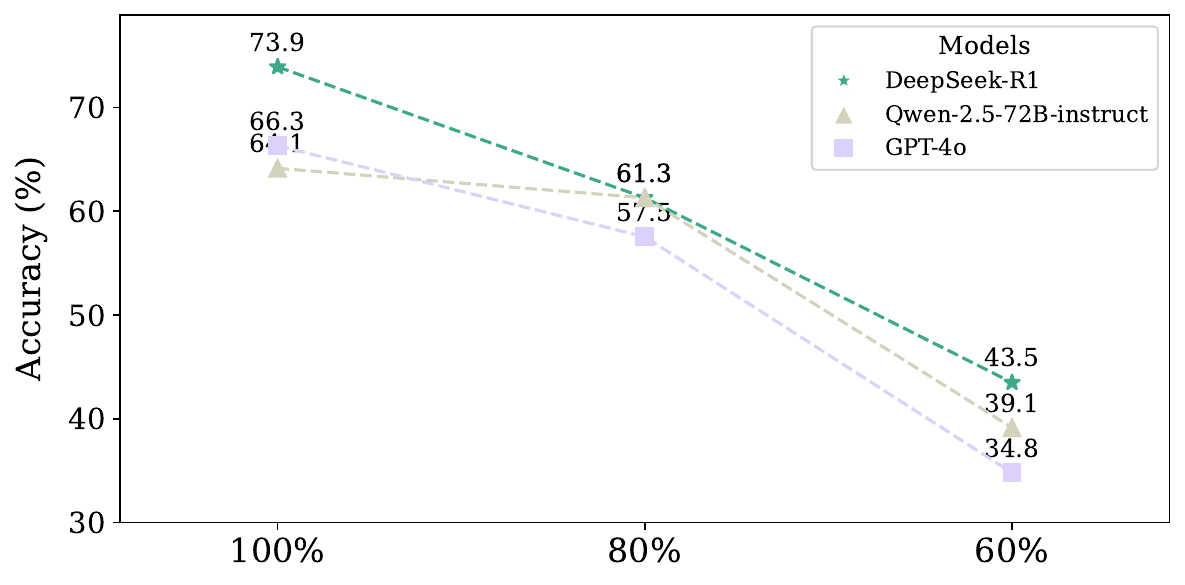}
    \caption{Accuracy of LLM Judges in Pair-wise Evaluation. The x-axis represents human labeling confidence, indicating the level of agreement among five annotators on the final labels.}
    \label{fig:llm_judge}
\end{figure}

We employ LLM-as-a-Judge for automatic evaluation within CharacterArena. To verify the reliability of this approach, we manually annotate 200 pairwise evaluation samples of multi-turn dialogues from consistent chat scenarios. Each sample is independently labeled by five human annotators, with the reference label determined via majority vote.

Figure~\ref{fig:llm_judge} presents a comparison of three LLM evaluators on the human-annotated test set. (1) DeepSeek-R1 attains 73.9\% accuracy on samples with full human consensus; (2) The accuracy of all LLM judges declines as the level of human agreement decreases, suggesting that models also struggle with ambiguous or contentious samples. Overall, statistical analysis indicates that LLM-based evaluations are generally consistent with human judgments. Based on its superior agreement with human annotations, DeepSeek-R1 is selected as the default Judge model in CharacterArena.

\section{Conclusion}
This paper proposes Comparative Policy Optimization (CPO), a simple yet effective reinforcement learning framework that addresses the challenge of reward ambiguity in open-ended role-playing dialogue. CPO replaces conventional sample-wise reward estimation with comparative group-wise scoring, aligning more closely with human evaluative behavior. To facilitate this, we introduce CharacterArena evaluation framework, which enables fairer trajectory-level comparisons under shared context. Experiments on CharacterEval, CharacterBench, and CharacterArena show that CPO outperforms existing RLHF methods. This work offers a novel and efficient pathway for reward modeling in subjective and creative tasks.
\section*{Limitations}

This work presents a new reward modeling method specifically designed for subjective, open-ended tasks, alongside an innovative framework for evaluating multi-turn dialogues. These contributions offer new avenues for optimizing and assessing open-domain tasks. Nevertheless, several limitations require further consideration. \textit{First,} our current optimization approach (e.g., CPO) primarily targets single-turn dialogue modeling. Future research will expand this to encompass the learning and refinement of multi-turn dialogue strategies. \textit{Second,} while this study has shown initial promise in complex role-playing scenarios, we intend to conduct more comprehensive and systematic evaluations across a broader spectrum of open-ended tasks, such as creative writing and story continuation.

\section*{Ethical Considerations}

This research utilized publicly available models such as LLaMA \cite{llama3}, Qwen \cite{qwen2.5}, CharacterEval Judge \cite{tu-etal-2024-charactereval}, CharacterBench Judge \cite{characterbench}, Doubao \cite{doubao}, Claude \cite{claude-3-5}, DeepSeek-R1 \cite{deepseekr1}, and GPT-4o \cite{gpt-4}, and toolkits like \texttt{LLaMA-Factory} \cite{zheng2024llamafactory} and \texttt{verl} \cite{verl}. All data used in this study are either publicly accessible online or synthetically generated by the aforementioned models. The primary language of focus in this work is Chinese. This work is intended solely for research purposes.

We adhered to strict ethical guidelines in our human evaluation. Five students from diverse backgrounds were recruited to participate. Before beginning the evaluation, participants received a clear and thorough explanation of the study's objectives, including any potential risks or disclaimers, and a detailed overview of the evaluation process itself. To ensure fair compensation and respect for their time, participants were paid 18 RMB per sample, a rate that exceeds the prevailing local labor compensation standard.

\bibliography{custom}
\clearpage
\onecolumn
\appendix
\section*{Appendix}

\startcontents[sections]
\printcontents[sections]{l}{1}{\setcounter{tocdepth}{2}}

\clearpage
\twocolumn

\section{Experimental Setup}
\label{appendix:implementation_details}

\subsection{Datasets}
\label{appendix:datasets}
The supervised fine-tuning (SFT) stage involves two fine-tuning tasks: role-playing multi-turn dialogue and story creation. Here, we provide a detailed introduction to the data sources used for these two tasks.

\paragraph{Role-Playing}
For SFT in role-playing tasks, we utilize the RoleplayPref dataset \cite{fft}, which originally contains 1,108 roles and 16,888 dialogues. Although the dataset was initially designed for preference learning, we repurpose it for SFT by extracting only the conversation histories and discarding the preferred and rejected candidate responses. To ensure the effectiveness of multi-turn dialogue training, we filter the dataset to retain only dialogues with at least five turns. This yields a final training set of 13,230 dialogues, with an average of 7.65 turns per dialogue and an average of 64.53 words per assistant response.

\paragraph{Story Creation}
The story creation task comprises Story Continuation Writing and Story Generation.  \textit{Story Continuation Writing} involves generating a continuation given a story prefix. We create a dataset by segmenting publicly available novels into prompts (initial sections) and target continuations (subsequent sections). This process yielded 50,000 samples, with an average prompt length of 629 words and an average target continuation length of 612 words.
\textit{Story Generation} requires generating a complete story given a premise. We use the GPT-WritingPrompts dataset\footnote{\href{https://huggingface.co/datasets/vkpriya/GPT-WritingPrompts}{https://huggingface.co/datasets/vkpriya/GPT-WritingPrompts}} for this purpose, which contains 5,000 samples with an average story length of approximately 400 words.

\subsection{Baselines}
\label{appendix:baselines}

To thoroughly validate the effectiveness of our method, we conduct a comprehensive comparison against a range of advanced models, including both open-source and closed-source options. 

\paragraph{Open-Source Models} Prior research has yielded numerous instruction-following models fine-tuned on role-playing dialogue datasets. However, the majority of these approaches do not explore reinforcement learning alignment techniques. To rigorously evaluate the performance of our method, we compare it against several mainstream Reinforcement Learning Fine-Tuning (RLFT) methods, including Rejection Sampling Fine-Tuning (\textbf{\textit{RFT}})\cite{rft}, Direct Preference Optimization (\textbf{\textit{DPO}})\cite{dpo}, and vanilla \textbf{\textit{GRPO}} \cite{DeepSeekMath}. We utilize two frequently used LLM backbones to demonstrate the robustness of our approach: the Qwen2.5 series (7B, 14B) \cite{qwen2.5}\footnote{\href{https://huggingface.co/Qwen/Qwen2.5-7B-Instruct}{https://huggingface.co/Qwen/Qwen2.5-7B-Instruct}} \footnote{\href{https://huggingface.co/Qwen/Qwen2.5-14B-Instruct}{https://huggingface.co/Qwen/Qwen2.5-14B-Instruct}} and LLaMA3-8B-Instruct \cite{llama3}\footnote{\href{https://huggingface.co/meta-llama/Meta-Llama-3-8B-Instruct}{https://huggingface.co/meta-llama/Meta-Llama-3-8B-Instruct}}. To ensure a fair comparison, all baseline models were trained using the same experimental configuration as our method. Furthermore, we also include a comparison against the performance of the advanced LLMs like Deepseek-R1\cite{deepseekr1}.

\paragraph{Closed-Source Models} Several closed-source models have emerged that specialize in role-playing. These include \textbf{\textit{Claude-3.7-sonnet}} \cite{claude-3-5}, \textbf{\textit{Doubao-PRO-Character}} \cite{doubao}, and \textbf{\textit{Minimax-abab5.5s}} \cite{minimax}. In addition to these, we also compare against \textbf{\textit{GPT-4o}} \cite{gpt-4} due to its widely recognized strong performance.

\subsection{Experimental Environments}
\label{appendix:experiments_environments}
All experiments are conducted on 8 NVIDIA A100-SXM4-80GB. Models are self-supervised fine-tuned with \texttt{LLaMA-Factory}~\citep{zheng2024llamafactory} \footnote{\href{https://github.com/hiyouga/LLaMA-Factory.git}{https://github.com/hiyouga/LLaMA-Factory.git}}, reinforcement-learning fine-tuned with \texttt{verl}~\citep{verl} \footnote{\href{https://github.com/volcengine/verl.git}{https://github.com/volcengine/verl.git}}, and inference is performed with \texttt{vLLM}~\citep{kwon2023efficient}. 

\subsection{Hyperparameters}
\label{appendix:hyperparameters}

\paragraph{SFT}
We employ the LoRA technique \cite{LoRA} in all SFT experiments, using a rank of 8 and an alpha value of 16 for the LoRA adapter applied to each linear module. For optimization, we utilize the AdamW optimizer \cite{adamw} with a learning rate of $5e-6$. A cosine learning rate scheduler is implemented, with a warm-up phase spanning $1\%$ of the total training steps. Training proceeded for $3$ epochs, using a batch size of 8 and gradient accumulation over $2$ steps.

\paragraph{RLFT}
For CPO and GRPO experiments, models are trained for $3$ epochs with a learning rate of $5e-7$ and a batch size of $128$. The KL coefficient  $\beta$ is set to $1e-3$. Following \cite{dapo}, $\epsilon_{high}$ and $\epsilon_{low}$ are set to $0.28$ and $0.2$, respectively. Considering the typical length of role-playing responses, we set the length control hyperparameter $L_{max}$ to $128$, $L_{cache}$ to $60$. During the rollout phase, we use a temperature of $1.0$ and a top-p value of $1.0$.

To generate preference data for RFT and DPO, we sample $N = 16$ responses per context using a temperature of $1.0$ and top-p of $1.0$. Reward scoring is performed by the Qwen2.5-72b-instruct model, using a temperature of $0.0$. The DPO hyperparameter $\beta$ is set to $0.1$. For CPO and GRPO experiments, models were trained for $3$ epochs with a learning rate of $5e-6$ and a per-device batch size of $8$.

\paragraph{Inference}
A temperature of $0.0$ is used for the LLM Judge to ensure more stable evaluation results.

\subsection{Evaluation Details}
\label{appendix:evaluation_details}

The effectiveness of the method is verified through experiments on two utterance-level automatic evaluation benchmarks (CharacterEval, CharacterBench), one session-level automatic evaluation benchmark (CharacterArena), and human evaluation.

\paragraph{CharacterEval} 
\cite{tu-etal-2024-charactereval} is a comprehensive Chinese benchmark for evaluating utterance-level role-playing capability. It features 1,785 multi-turn dialogues, 4,564 test examples, and 77 characters from Chinese novels and scripts. Evaluation is conducted using CharacterRM \footnote{\href{https://huggingface.co/morecry/BaichuanCharRM}{https://huggingface.co/morecry/BaichuanCharRM}} and employs multifaceted metrics across three key aspects: \textbf{\textit{Conversational Ability}} (fluency, coherency, consistency), \textbf{\textit{Character Consistency}} (knowledge: exposure, accuracy, hallucination; persona: behavior, utterance), and \textbf{\textit{Role-playing Attractiveness}} (human-likeness, communication skills, expression diversity, empathy).

\paragraph{CharacterBench}
\cite{characterbench} is a large bilingual (zh/en) generative benchmark comprising 22,859 human-annotated samples that cover 3,956 characters across 25 detailed character categories. This work focuses exclusively on the Chinese data within CharacterBench. Evaluation is conducted using CharacterJudge \footnote{\href{https://huggingface.co/thu-coai/CharacterJudge}{https://huggingface.co/thu-coai/CharacterJudge}} across 11 dimensions categorized under 6 key aspects: \textbf{\textit{Memory, Knowledge, Persona, Emotion, Morality, and Believability.}} Specifically, these dimensions include  Memory Consistency, Fact Accuracy, Boundary Consistency, Attribute Consistency (Bot), Attribute Consistency (Human), Behavior Consistency (Bot), Behavior Consistency (Human), Emotional Self-regulation, Empathetic Responsiveness, Morality Stability, Morality Robustness, Human-likeness, and Engagement. The benchmark differentiates between sparse and dense dimensions, indicating the consistent presence or absence of specific character features within generated responses.

\paragraph{CharacterArena} 
The aforementioned benchmarks are utterance-level, which primarily evaluate models based on the provided conversation history. Consequently, they may not fully capture a model's capability in multi-turn role-playing dialogues due to potential biases arising from not being self-generated by the model. To mitigate this limitation, we introduce \textbf{CharacterArena} (Section \ref{sec:characterarena}), a dynamic, session-level evaluation. CharacterArena works by comparing dialogue segments generated by two different models within the same scenario. This comparative framework transforms the subjective task of evaluating role-playing quality into a more objective ranking of generated dialogue snippets. Our evaluation considers several dimensions: dialogue attractiveness (\textit{our primary focus}), multi-turn dialogue coherence (encompassing both logical flow and linguistic consistency), and character persona consistency (\textit{as an auxiliary measure}). The evaluation is conducted using DeepSeek-R1 \cite{deepseekr1} as the LLM Judge model. For each pair of models, we run 50 contextualized multi-turn role-playing simulations, each lasting for 15 turns. Dialogue order is randomized to prevent position bias. See Appendix \ref{apx:eval_prompts} for detailed evaluation prompts.

\paragraph{Human Evaluation}
To further enhance the comprehensiveness and reliability of our evaluation, we supplement automatic assessments with human evaluations. Human annotators are presented with two dialogues (A and B), generated by different models under the same role-playing settings as CharacterArena. They are asked to select one of three options: “A wins”, “Tie”, or “B wins”. To ensure fairness, we randomize the order of dialogues to eliminate position bias. Each model pair is evaluated on 50 contextualized multi-turn role-playing simulations, with each simulation independently assessed by three human annotators. If all three annotators provide completely different outcomes, the sample is considered invalid and excluded.

\subsection{The Design of our Evaluation Criteria}

To capture the multifaceted nature of high-quality role-playing dialogue, our evaluation framework centers on three carefully defined dimensions: \textbf{Creativity}, \textbf{Coherence}, and \textbf{Consistency}.

\paragraph{Creativity}
Creativity is the central goal in entertainment-driven role-playing systems, as it directly influences how engaging and dynamic the interaction becomes. Our evaluation considers three key aspects of creativity: (1) \textit{Plot Development} — the ability to advance the storyline and generate rich narrative branches; (2) \textit{Immersion} — the extent to which the dialogue and narrative can captivate users and evoke a sense of presence in the fictional world; (3) \textit{Narrative Skill} — the use of expressive language and stylistic techniques to enhance storytelling appeal.

The Creativity dimension assesses whether the model can create compelling plots, rather than simply producing novel language. We expect the model to proactively shape the storyline within the constraints of the character's persona, adding dramatic tension and increasing the interactive playability of the dialogue.

\paragraph{Coherence}
We evaluate dialogue coherence across three hierarchical levels: (1) \textit{Utterance-level}: assessing the fluency and naturalness of individual responses; (2) \textit{Conversation-level}: evaluating logical continuity and contextual relevance across multiple turns; (3) \textit{World-level}: examining the internal consistency of the fictional setting, including temporal and spatial coherence.

Coherence becomes particularly challenging in extended interactions. While achieving fluency at the utterance level is relatively straightforward, maintaining conversation-level logic requires strong memory and reasoning capabilities. For instance, a story set in a medieval fantasy world should not suddenly introduce modern technology, nor should characters behave in ways inconsistent with their era or background. To comprehensively assess dialogue quality, we emphasize coherence across linguistic, logical, and world-building dimensions.

\paragraph{Consistency} 
The Consistency dimension focuses on the sustained fidelity to a character's persona. Even in the face of disruptive or provocative user inputs, the model should maintain the integrity of its role, avoiding character drift or narrative inconsistency. This dimension contrasts with Creativity: while creativity encourages the model to expand on the character's profile with new plot details, consistency demands that such expansions remain aligned with the character's core attributes.

Together, these dimensions form a balanced framework: Consistency ensures believability, Creativity drives engagement, and Coherence bridges the two by enabling interactions that are logically sound and narratively fluid.

\section{Additional Experimental Results}
\label{apx:add_experiments}
\subsection{Benchmark Results}
\begin{figure}[th]
    \centering
    \begin{subfigure}[b]{\linewidth}
        \centering
        \includegraphics[width=0.9\textwidth]{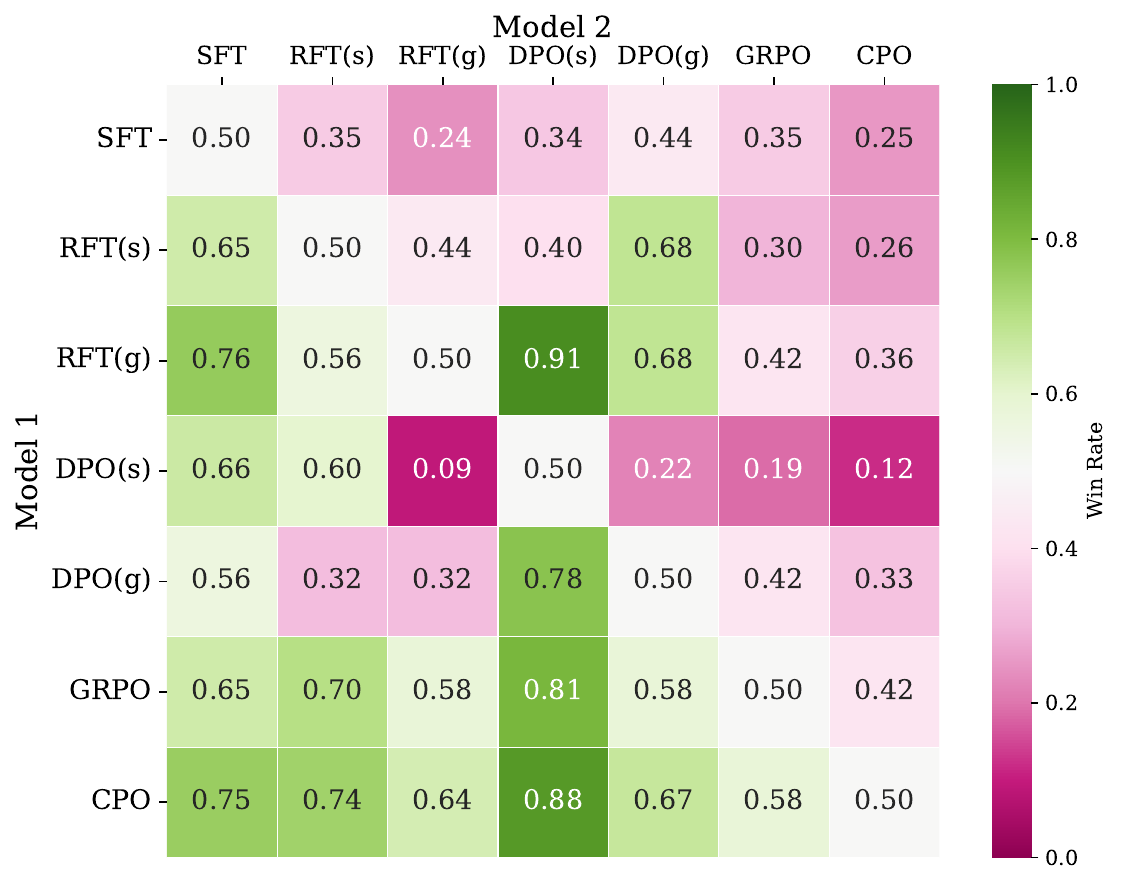}
        \caption{Qwen-2.5-7b-Instruct}
        \label{fig:win_rate_qwen_7b_full}
    \end{subfigure}
    \hfill
    \begin{subfigure}[b]{\linewidth}
        \centering
        \includegraphics[width=0.9\textwidth]{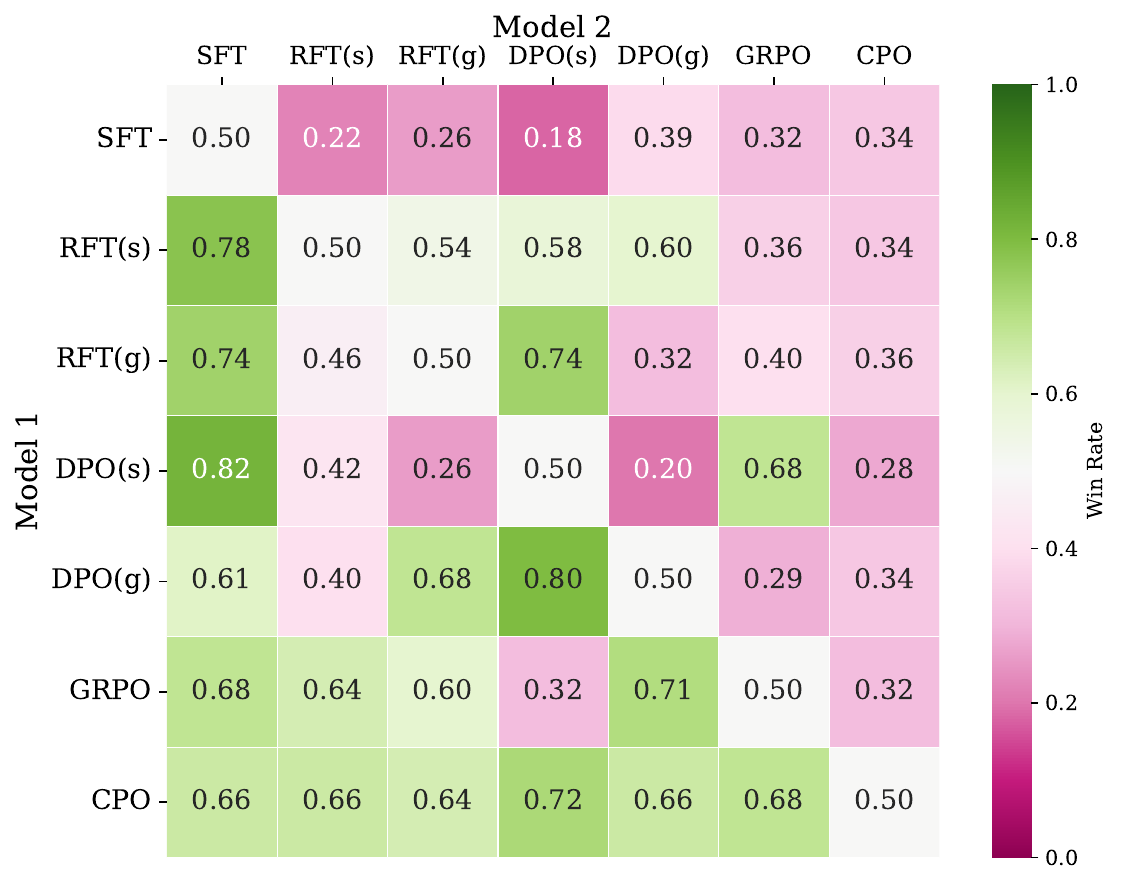}
        \caption{LLaMA-3-8b-Instruct}
        \label{fig:win_rate_llama_full}
    \end{subfigure}
    \caption{Win Rate Matrices on the CharacterArena Benchmark. Values in cell $(i,j)$ indicate the preference rate of Model $i$ over Model $j$.}
    \label{fig:win rate full}
\end{figure}

We present the complete benchmark evaluation results in this section. Detailed outcomes for CharacterEval are provided in Table~\ref{tab:character_eval_full}, those for CharacterBench (zh) in Table~\ref{tab:character_bench_zh_full}, and the win rate matrices for CharacterArena in Figure~\ref{fig:win rate full}. Our experiments show that the proposed \textit{group-wise reward modeling }approach consistently enhances performance across all alignment methods, including RFT, DPO, and CPO.

Notably, DPO outperforms online reinforcement learning methods (CPO and GRPO) on both CharacterEval and CharacterBench. We observe that models trained with DPO tend to produce longer responses, with an average output length reaching the preset upper limit of 128 tokens. This verbosity appears beneficial under the LLM-as-a-Judge evaluation protocol, potentially inflating scores and exposing a systemic bias in current evaluation frameworks that favor longer outputs.

\subsection{Additional Analysis}
\subsubsection{Response Length Analysis}

\begin{figure}[t]
    \centering
    \begin{subfigure}[b]{\linewidth}
        \centering
        \includegraphics[width=\textwidth]{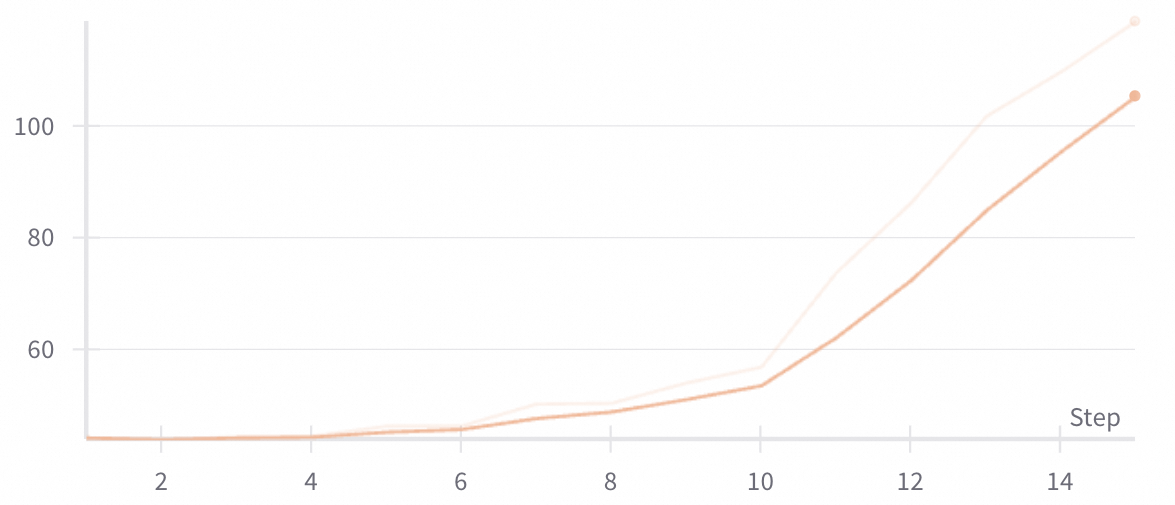}
        \caption{Without Length Control}
        \label{fig:wo_length_control} 
    \end{subfigure}
    \begin{subfigure}[b]{\linewidth}
        \centering
        \includegraphics[width=\textwidth]{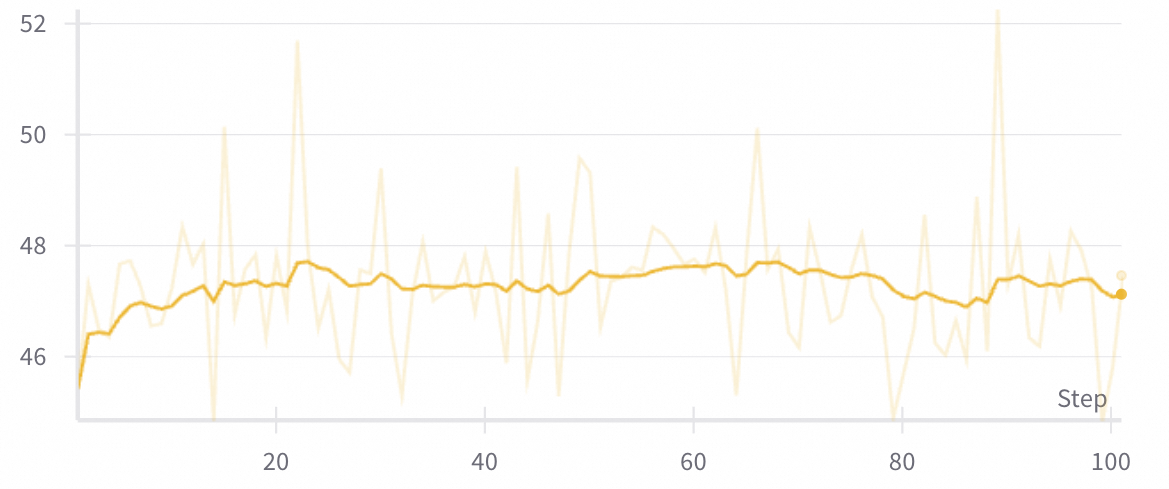}
        \caption{With Length Control}
        \label{fig:with_length_control} 
    \end{subfigure}
    \caption{Average Response Length during Training.} 
    \label{fig:average_response_length_during_training} 
\end{figure}

As mentioned in Section \ref{sec:cpo}, we employ a \textit{soft overlength penalty} to mitigate the generation of excessively long responses during training. Figure \ref{fig:average_response_length_during_training} illustrates the change in average response length during training, both with and without length control. It can be observed that without length control, the average response length quickly reaches the predefined upper limit within a few steps. In contrast, when length control is applied, the average response length remains within a controllable range throughout the training process. This intervention is crucial, as evaluations based on the LLM-as-a-Judge protocol are often vulnerable to length bias, which can lead to reward hacking.

\begin{table*}[ht]
\centering
\setstretch{1.25}
\resizebox{\textwidth}{!}{
\begin{tabular}{lcccccccccccccccc}
\toprule[1.2pt]
\multicolumn{1}{c|}{} &
  \multicolumn{4}{c|}{\textit{\textbf{Conversational Ability}}} &
  \multicolumn{6}{c|}{\textit{\textbf{Character Consistency}}} &
  \multicolumn{5}{c|}{\textit{\textbf{Role-playing Attractiveness}}} &
   \\
\multicolumn{1}{c|}{\multirow{-2}{*}{\textit{\textbf{Model}}}} &
  \textit{\textbf{Flu.}} &
  \textit{\textbf{Coh.}} &
  \textit{\textbf{Con.}} &
  \multicolumn{1}{c|}{\textit{\textbf{Avg.}}} &
  \textit{\textbf{Exp.}} &
  \textit{\textbf{Acc.}} &
  \textit{\textbf{Hall.}} &
  \textit{\textbf{Beh.}} &
  \textit{\textbf{Utt.}} &
  \multicolumn{1}{c|}{\textit{\textbf{Avg.}}} &
  \textit{\textbf{Hum.}} &
  \textit{\textbf{Com.}} &
  \textit{\textbf{Div.}} &
  \textit{\textbf{Emp.}} &
  \multicolumn{1}{c|}{\textit{\textbf{Avg.}}} &
  \multirow{-2}{*}{\textit{\textbf{Avg.}}} \\ \midrule
\rowcolor[HTML]{DFEBF6} 
\multicolumn{17}{l}{\cellcolor[HTML]{DFEBF6}\textit{\textbf{Open-source LLMs}}} \\ 
\rowcolor[HTML]{EFEFEF} 
\multicolumn{1}{l|}{\cellcolor[HTML]{EFEFEF}Qwen-2.5-7b-SFT} &
  3.51 &
  3.92 &
  \textbf{3.71} &
  \multicolumn{1}{c|}{\cellcolor[HTML]{EFEFEF}3.71} &
  2.15 &
  2.98 &
  2.97 &
  3.52 &
  3.10 &
  \multicolumn{1}{c|}{\cellcolor[HTML]{EFEFEF}2.94} &
  \textbf{3.57} &
  3.23 &
  2.90 &
  3.12 &
  \multicolumn{1}{c|}{\cellcolor[HTML]{EFEFEF}3.20} &
  3.29 \\
\multicolumn{1}{l|}{+ RFT (s)} &
  3.52 &
  3.91 &
  3.60 &
  \multicolumn{1}{c|}{3.68} &
  2.20 &
  2.94 &
  2.94 &
  3.70 &
  3.06 &
  \multicolumn{1}{c|}{2.97} &
  3.41 &
  3.24 &
  3.10 &
  3.10 &
  \multicolumn{1}{c|}{3.21} &
  3.29 \\
\multicolumn{1}{l|}{+ RFT (g)} &
  3.50 &
  3.88 &
  3.55 &
  \multicolumn{1}{c|}{3.65} &
  2.28 &
  2.95 &
  2.94 &
  3.75 &
  3.08 &
  \multicolumn{1}{c|}{3.00} &
  3.38 &
  3.28 &
  3.25 &
  3.06 &
  \multicolumn{1}{c|}{3.24} &
  3.30 \\ \cdashline{1-17}
\multicolumn{1}{l|}{+ DPO (s)} &
  3.26 &
  3.62 &
  3.14 &
  \multicolumn{1}{c|}{3.34} &
  \textbf{2.87} &
  \textbf{3.13} &
  3.01 &
  \textbf{3.97} &
  2.99 &
  \multicolumn{1}{c|}{3.19} &
  2.90 &
  3.56 &
  \textbf{3.75} &
  3.19 &
  \multicolumn{1}{c|}{3.35} &
  3.29 \\
\multicolumn{1}{l|}{+ DPO (g)} &
  3.57 &
  3.96 &
  3.58 &
  \multicolumn{1}{c|}{3.70} &
  2.80 &
  3.12 &
  \textbf{3.19} &
  3.92 &
  3.20 &
  \multicolumn{1}{c|}{\textbf{3.24}} &
  3.21 &
  \textbf{3.76} &
  3.52 &
  \textbf{3.41} &
  \multicolumn{1}{c|}{\textbf{3.48}} &
  \textbf{3.47} \\ \cdashline{1-17}
\multicolumn{1}{l|}{+ GRPO} &
  3.54 &
  3.96 &
  \textbf{3.71} &
  \multicolumn{1}{c|}{3.73} &
  2.17 &
  2.99 &
  2.96 &
  3.55 &
  3.10 &
  \multicolumn{1}{c|}{2.95} &
  3.54 &
  3.25 &
  2.96 &
  3.13 &
  \multicolumn{1}{c|}{3.22} &
  3.30 \\
\multicolumn{1}{l|}{+ CPO} &
  \textbf{3.58} &
  \textbf{3.97} &
  \textbf{3.71} &
  \multicolumn{1}{c|}{\textbf{3.75}} &
  2.29 &
  3.01 &
  3.03 &
  3.67 &
  \textbf{3.13} &
  \multicolumn{1}{c|}{3.03} &
  3.49 &
  3.40 &
  3.07 &
  3.23 &
  \multicolumn{1}{c|}{3.30} &
  3.36 \\ \hline
\rowcolor[HTML]{EFEFEF} 
\multicolumn{1}{l|}{\cellcolor[HTML]{EFEFEF}LLaMA-3-8b-SFT} &
  \textbf{3.38} &
  3.80 &
  \textbf{3.48} &
  \multicolumn{1}{c|}{\cellcolor[HTML]{EFEFEF}3.55} &
  2.19 &
  2.90 &
  2.85 &
  3.64 &
  3.00 &
  \multicolumn{1}{c|}{\cellcolor[HTML]{EFEFEF}2.92} &
  3.24 &
  3.21 &
  3.06 &
  3.02 &
  \multicolumn{1}{c|}{\cellcolor[HTML]{EFEFEF}3.13} &
  3.20 \\
\multicolumn{1}{l|}{+ RFT (s)} &
  3.19 &
  3.61 &
  3.19 &
  \multicolumn{1}{c|}{3.33} &
  2.28 &
  2.86 &
  2.73 &
  3.67 &
  2.83 &
  \multicolumn{1}{c|}{2.87} &
  2.98 &
  3.15 &
  3.14 &
  2.93 &
  \multicolumn{1}{c|}{3.05} &
  3.08 \\
\multicolumn{1}{l|}{+ RFT (g)} &
  3.14 &
  3.55 &
  3.08 &
  \multicolumn{1}{c|}{3.25} &
  2.35 &
  2.85 &
  2.71 &
  3.64 &
  2.78 &
  \multicolumn{1}{l|}{2.87} &
  2.93 &
  3.16 &
  3.14 &
  2.96 &
  \multicolumn{1}{c|}{3.05} &
  3.06 \\ \cdashline{1-17}
\multicolumn{1}{l|}{+ DPO (s)} &
  3.53 &
  3.91 &
  3.47 &
  \multicolumn{1}{c|}{\textbf{3.63}} &
  2.64 &
  3.01 &
  3.04 &
  4.15 &
  3.16 &
  \multicolumn{1}{l|}{3.20} &
  3.19 &
  3.60 &
  3.89 &
  3.24 &
  \multicolumn{1}{c|}{3.48} &
  3.44 \\
\multicolumn{1}{l|}{+ DPO (g)} &
  3.53 &
  3.91 &
  3.46 &
  \multicolumn{1}{c|}{\textbf{3.63}} &
  \textbf{2.82} &
  \textbf{3.06} &
  \textbf{3.12} &
  \textbf{4.17} &
  \textbf{3.26} &
  \multicolumn{1}{l|}{\textbf{3.29}} &
  3.16 &
  \textbf{3.70} &
  \textbf{3.98} &
  \textbf{3.32} &
  \multicolumn{1}{c|}{\textbf{3.54}} &
  \textbf{3.49} \\ \cdashline{1-17}
\multicolumn{1}{l|}{+ GRPO} &
  3.35 &
  3.79 &
  \textbf{3.48} &
  \multicolumn{1}{c|}{3.54} &
  2.1 &
  2.87 &
  2.82 &
  3.52 &
  2.98 &
  \multicolumn{1}{c|}{2.86} &
  \textbf{3.32} &
  3.08 &
  2.92 &
  2.97 &
  \multicolumn{1}{c|}{3.07} &
  3.16 \\
\multicolumn{1}{l|}{+ CPO} &
  3.36 &
  \textbf{3.81} &
  \textbf{3.48} &
  \multicolumn{1}{c|}{3.55} &
  2.19 &
  2.90 &
  2.87 &
  3.62 &
  2.99 &
  \multicolumn{1}{c|}{2.91} &
  3.26 &
  3.20 &
  3.02 &
  3.04 &
  \multicolumn{1}{c|}{3.13} &
  3.20 \\ \midrule \midrule
\rowcolor[HTML]{DFEBF6} 
\multicolumn{17}{l}{\cellcolor[HTML]{DFEBF6}\textit{\textbf{Close-source LLMs}}} \\ 
\multicolumn{1}{l|}{MiniMax-abab5.5s} &
  3.61 &
  3.93 &
  3.81 &
  \multicolumn{1}{c|}{3.78} &
  1.84 &
  2.91 &
  2.94 &
  2.77 &
  3.13 &
  \multicolumn{1}{c|}{2.72} &
  3.77 &
  2.67 &
  2.15 &
  3.01 &
  \multicolumn{1}{c|}{2.90} &
  3.13 \\
\multicolumn{1}{l|}{Deepseek-R1} &
  3.53 &
  3.83 &
  3.96 &
  \multicolumn{1}{c|}{3.77} &
  1.34 &
  3.07 &
  2.88 &
  1.70 &
  3.21 &
  \multicolumn{1}{c|}{2.44} &
  \textbf{4.38} &
  1.71 &
  1.56 &
  2.93 &
  \multicolumn{1}{c|}{2.64} &
  2.95 \\
\multicolumn{1}{l|}{GPT-4o} &
  3.54 &
  3.89 &
  3.47 &
  \multicolumn{1}{c|}{3.63} &
  2.58 &
  3.13 &
  2.99 &
  2.83 &
  2.98 &
  \multicolumn{1}{c|}{2.90} &
  3.17 &
  3.54 &
  2.20 &
  3.32 &
  \multicolumn{1}{c|}{3.06} &
  3.20 \\
\multicolumn{1}{l|}{Doubao-Pro-Character} &
  3.61 &
  3.94 &
  3.64 &
  \multicolumn{1}{c|}{3.73} &
  \textbf{2.85} &
  \textbf{3.41} &
  \textbf{3.17} &
  3.87 &
  3.17 &
  \multicolumn{1}{c|}{\textbf{3.29}} &
  3.42 &
  \textbf{3.74} &
  3.35 &
  \textbf{3.53} &
  \multicolumn{1}{c|}{\textbf{3.51}} &
  \textbf{3.51} \\
\multicolumn{1}{l|}{Claude-3.7-sonnet} &
  \textbf{3.71} &
  \textbf{3.99} &
  \textbf{4.00} &
  \multicolumn{1}{c|}{\textbf{3.90}} &
  2.03 &
  3.06 &
  3.04 &
  \textbf{3.91} &
  \textbf{3.26} &
  \multicolumn{1}{c|}{3.06} &
  3.91 &
  2.75 &
  \textbf{3.38} &
  2.97 &
  \multicolumn{1}{c|}{3.25} &
  3.41 \\ \bottomrule[1.2pt]
\end{tabular}%
}
\caption{The Overall Results on \textbf{CharacterEval} Benchmark.`\textit{g}' and `\textit{s}' denote group-wise and sample-wise rewarding, respectively.}
\label{tab:character_eval_full}
\end{table*}
\begin{table*}[]
\centering
\setstretch{1.25}
\resizebox{\textwidth}{!}{%
\begin{tabular}{lcccccccccccccc}
\toprule[1.2pt]
    \multicolumn{15}{l}{\makecell[l]{
    \textbb{MC}: Memory Consistency ~~~ 
    \textbb{FA}: Fact Accuracy ~~~ 
    \textbb{BC$_K$}: Boundary Consistency ~~~ 
    \textbb{AC$^h$}: Attribute Consistency (Human) ~~~
    \textbb{EG}: Engagement \\
    \textbb{AC$^b$}: Attribute Consistency (Bot)~~~
    \textbb{BC$_P^b$}: Behavior Consistency (Bot) ~~~ 
    \textbb{BC$_P^h$}: Behavior Consistency (Human) ~~~
    \textbb{HL}: Human-likeness 
    \\
    \textbb{ES}: Emotional Self-regulation ~~~
    \textbb{ER}: Empathetic Responsiveness ~~~
    \textbb{MS}: Morality Stability ~~~ 
    \textbb{MR}: Morality Robustness
    }} \\
\hline
\multicolumn{1}{c|}{} & \multicolumn{1}{c|}{} & \multicolumn{1}{c|}{\textbf{Memory}} & \multicolumn{2}{c|}{\textbf{Knowledge}} & \multicolumn{4}{c|}{\textbf{Person}} & \multicolumn{2}{c|}{\textbf{Emotion}} & \multicolumn{2}{c|}{\textbf{Morality}} & \multicolumn{2}{c}{\textbf{Believability}} \\
\multicolumn{1}{c|}{\multirow{-2}{*}{\textbf{Model}}} & \multicolumn{1}{c|}{\multirow{-2}{*}{\textbf{Average}}} & \multicolumn{1}{c|}{\textbf{\textbb{MC}}} & \textbf{\textbb{FA}} & \multicolumn{1}{c|}{\textbf{\textbb{BC$_K$}}} & \textbf{\textbb{AC$^b$}} & \textbf{\textbb{AC$^h$}} & \textbf{\textbb{BC$_P^b$}} & \multicolumn{1}{c|}{\textbf{\textbb{BC$_P^h$}}} & \textbf{\textbb{ES}} & \multicolumn{1}{c|}{\textbf{\textbb{ER}}} & \textbf{\textbb{MS}} & \multicolumn{1}{c|}{\textbf{\textbb{MR}}} & \textbf{\textbb{HL}} & \textbf{\textbb{EG}} \\ \midrule
\rowcolor[HTML]{DFEBF6} 
\multicolumn{15}{l}{\cellcolor[HTML]{DFEBF6}\textit{\textbf{Open-source LLMs}}} \\
\rowcolor[HTML]{EFEFEF} 
\multicolumn{1}{l|}{\cellcolor[HTML]{EFEFEF}Qwen-2.5-7b-SFT} & \multicolumn{1}{c|}{\cellcolor[HTML]{EFEFEF}3.33} & \multicolumn{1}{c|}{\cellcolor[HTML]{EFEFEF}3.15} & 2.32 & \multicolumn{1}{c|}{\cellcolor[HTML]{EFEFEF}3.59} & 3.33 & 3.51 & 3.17 & \multicolumn{1}{c|}{\cellcolor[HTML]{EFEFEF}3.12} & 3.04 & \multicolumn{1}{c|}{\cellcolor[HTML]{EFEFEF}2.68} & 4.68 & \multicolumn{1}{c|}{\cellcolor[HTML]{EFEFEF}4.76} & 2.84 & 3.13 \\
\multicolumn{1}{l|}{+ RFT(s)} & \multicolumn{1}{c|}{3.44} & \multicolumn{1}{c|}{3.09} & 2.43 & \multicolumn{1}{c|}{3.61} & 3.67 & 3.85 & 3.39 & \multicolumn{1}{c|}{3.20} & \textbf{3.15} & \multicolumn{1}{c|}{2.83} & 4.74 & \multicolumn{1}{c|}{4.75} & 2.92 & \textbf{3.14} \\
\multicolumn{1}{l|}{+ RFT(g)} & \multicolumn{1}{c|}{3.51} & \multicolumn{1}{c|}{3.27} & 2.52 & \multicolumn{1}{c|}{3.73} & 3.90 & 3.81 & 3.47 & \multicolumn{1}{c|}{3.11} & 3.10 & \multicolumn{1}{c|}{2.87} & 4.82 & \multicolumn{1}{c|}{4.75} & \textbf{3.06} & 3.20 \\ \cdashline{1-15}
\multicolumn{1}{l|}{+ DPO(s)} & \multicolumn{1}{c|}{3.47} & \multicolumn{1}{c|}{3.56} & \textbf{2.59} & \multicolumn{1}{c|}{3.55} & 3.81 & \textbf{3.92} & 3.37 & \multicolumn{1}{c|}{\textbf{3.28}} & 3.08 & \multicolumn{1}{c|}{2.79} & 4.72 & \multicolumn{1}{c|}{4.85} & 2.54 & 3.09 \\
\multicolumn{1}{l|}{+ DPO(g)} & \multicolumn{1}{c|}{\textbf{3.54}} & \multicolumn{1}{c|}{\textbf{3.71}} & 2.40 & \multicolumn{1}{c|}{3.56} & \textbf{4.05} & \textbf{3.92} & \textbf{3.54} & \multicolumn{1}{c|}{3.11} & 3.30 & \multicolumn{1}{c|}{\textbf{3.09}} & 4.90 & \multicolumn{1}{c|}{\textbf{4.96}} & 2.69 & 2.87 \\ \cdashline{1-15}
\multicolumn{1}{l|}{+ GRPO} & \multicolumn{1}{c|}{3.41} & \multicolumn{1}{c|}{3.29} & 2.42 & \multicolumn{1}{c|}{3.74} & 3.56 & 3.69 & 3.14 & \multicolumn{1}{c|}{3.14} & 3.03 & \multicolumn{1}{c|}{2.73} & 4.76 & \multicolumn{1}{c|}{4.80} & 2.94 & 3.12 \\
\multicolumn{1}{l|}{+ CPO} & \multicolumn{1}{c|}{3.49} & \multicolumn{1}{c|}{3.44} & 2.48 & \multicolumn{1}{c|}{\textbf{3.78}} & 3.75 & 3.91 & 3.38 & \multicolumn{1}{c|}{3.11} & 3.09 & \multicolumn{1}{c|}{2.81} & \textbf{4.91} & \multicolumn{1}{c|}{4.77} & 2.84 & 3.11 \\ \hline
\rowcolor[HTML]{EFEFEF} 
\multicolumn{1}{l|}{\cellcolor[HTML]{EFEFEF}LLaMA-3-8b-SFT} & \multicolumn{1}{c|}{\cellcolor[HTML]{EFEFEF}3.29} & \multicolumn{1}{c|}{\cellcolor[HTML]{EFEFEF}3.10} & 2.24 & \multicolumn{1}{c|}{\cellcolor[HTML]{EFEFEF}3.58} & 3.43 & 3.58 & 3.17 & \multicolumn{1}{c|}{\cellcolor[HTML]{EFEFEF}3.15} & 2.86 & \multicolumn{1}{c|}{\cellcolor[HTML]{EFEFEF}2.70} & 4.64 & \multicolumn{1}{c|}{\cellcolor[HTML]{EFEFEF}4.63} & 2.73 & 2.92 \\
\multicolumn{1}{l|}{+ RFT(s)} & \multicolumn{1}{c|}{3.27} & \multicolumn{1}{c|}{3.21} & 2.24 & \multicolumn{1}{c|}{3.25} & 3.60 & 3.50 & 3.12 & \multicolumn{1}{c|}{3.07} & 2.96 & \multicolumn{1}{c|}{2.67} & 4.61 & \multicolumn{1}{c|}{4.55} & 2.91 & 2.87 \\
\multicolumn{1}{l|}{+ RFT(g)} & \multicolumn{1}{c|}{3.21} & \multicolumn{1}{c|}{3.10} & 2.19 & \multicolumn{1}{c|}{3.35} & 3.47 & 3.49 & 3.13 & \multicolumn{1}{c|}{3.03} & 2.79 & \multicolumn{1}{c|}{2.59} & 4.47 & \multicolumn{1}{c|}{4.51} & 2.68 & 2.92 \\ \cdashline{1-15}
\multicolumn{1}{l|}{+ DPO(s)} & \multicolumn{1}{c|}{\textbf{3.54}} & \multicolumn{1}{c|}{\textbf{3.75}} & 2.31 & \multicolumn{1}{c|}{3.63} & 4.01 & \textbf{4.15} & \textbf{3.54} & \multicolumn{1}{c|}{3.30} & \textbf{3.08} & \multicolumn{1}{c|}{2.86} & \textbf{4.80} & \multicolumn{1}{c|}{4.70} & 2.87 & \textbf{3.06} \\
\multicolumn{1}{l|}{+ DPO(g)} & \multicolumn{1}{c|}{\textbf{3.54}} & \multicolumn{1}{c|}{3.73} & \textbf{2.35} & \multicolumn{1}{c|}{3.65} & \textbf{4.15} & 4.10 & 3.49 & \multicolumn{1}{c|}{\textbf{3.36}} & 3.17 & \multicolumn{1}{c|}{\textbf{2.87}} & 4.68 & \multicolumn{1}{c|}{4.72} & 2.82 & 2.95 \\ \cdashline{1-15}
\multicolumn{1}{l|}{+ GRPO} & \multicolumn{1}{c|}{3.31} & \multicolumn{1}{c|}{3.19} & 2.23 & \multicolumn{1}{c|}{3.68} & 3.60 & 3.55 & 3.23 & \multicolumn{1}{c|}{3.04} & 2.85 & \multicolumn{1}{c|}{2.66} & 4.67 & \multicolumn{1}{c|}{4.65} & 2.73 & 2.98 \\
\multicolumn{1}{l|}{+ CPO} & \multicolumn{1}{c|}{3.36} & \multicolumn{1}{c|}{3.17} & 2.23 & \multicolumn{1}{c|}{\textbf{3.82}} & 3.54 & 3.63 & 3.21 & \multicolumn{1}{c|}{3.21} & 2.90 & \multicolumn{1}{c|}{2.68} & 4.60 & \multicolumn{1}{c|}{\textbf{4.76}} & \textbf{2.92} & 3.02 \\ \midrule \midrule
\rowcolor[HTML]{DFEBF6} 
\multicolumn{15}{l}{\cellcolor[HTML]{DFEBF6}\textit{\textbf{Closed-source LLMs}}} \\
\multicolumn{1}{l|}{MiniMax-abab5.5s} & \multicolumn{1}{c|}{3.52} & \multicolumn{1}{c|}{3.76} & 2.76 & \multicolumn{1}{c|}{3.45} & 4.18 & 4.02 & 3.35 & \multicolumn{1}{c|}{3.04} & 3.04 & \multicolumn{1}{c|}{2.71} & 4.69 & \multicolumn{1}{c|}{4.65} & 3.02 & 3.15 \\
\multicolumn{1}{l|}{Deepseek-R1} & \multicolumn{1}{c|}{3.77} & \multicolumn{1}{c|}{3.56} & 3.17 & \multicolumn{1}{c|}{3.76} & 4.28 & 4.21 & 3.81 & \multicolumn{1}{c|}{4.00} & 3.17 & \multicolumn{1}{c|}{3.08} & 4.69 & \multicolumn{1}{c|}{4.53} & 3.20 & 3.49 \\
\multicolumn{1}{l|}{GPT-4o} & \multicolumn{1}{c|}{3.86} & \multicolumn{1}{c|}{3.83} & \textbf{3.28} & \multicolumn{1}{c|}{3.86} & \textbf{4.73} & 4.38 & 3.81 & \multicolumn{1}{c|}{3.60} & 3.53 & \multicolumn{1}{c|}{3.51} & \textbf{4.96} & \multicolumn{1}{c|}{\textbf{4.91}} & 2.68 & 3.06 \\
\multicolumn{1}{l|}{Doubao-Pro-Character} & \multicolumn{1}{c|}{3.90} & \multicolumn{1}{c|}{3.67} & 3.04 & \multicolumn{1}{c|}{3.98} & 4.49 & 4.53 & 3.88 & \multicolumn{1}{c|}{3.92} & 3.31 & \multicolumn{1}{c|}{3.17} & 4.91 & \multicolumn{1}{c|}{\textbf{4.91}} & \textbf{3.40} & 3.49 \\
\multicolumn{1}{l|}{Claude-3.7-sonnet} & \multicolumn{1}{c|}{\textbf{4.07}} & \multicolumn{1}{c|}{\textbf{4.03}} & 3.12 & \multicolumn{1}{c|}{\textbf{4.21}} & \textbf{4.73} & \textbf{4.60} & \textbf{4.03} & \multicolumn{1}{c|}{\textbf{4.34}} & \textbf{3.84} & \multicolumn{1}{c|}{\textbf{3.61}} & 4.94 & \multicolumn{1}{c|}{4.83} & 3.14 & \textbf{3.52} \\ \bottomrule[1.2pt]
\end{tabular}%
}
\caption{The Overall Results on \textbf{CharacterBench (zh)} Benchmark. `\textit{g}' and `\textit{s}' denote group-wise and sample-wise rewarding, respectively.}
\label{tab:character_bench_zh_full}
\end{table*}

\newpage
\onecolumn

\section{Others}

\subsection{Example of Role Profile}
\label{apx: Example of Role Profile}
\mycolorbox{The Example of Chat circumstances (zh)}
{
\footnotesize
\setstretch{1.2}
\begin{CJK}{UTF8}{gkai}

\textbf{角色:} 孙悟空

\textbf{角色背景: }

你是孙悟空，石头化生的灵明石猴。你曾在花果山为众猴之王，后拜须菩提祖师学得七十二变等神通。你性格桀骜不驯，嫉恶如仇，渴望自由，蔑视权威。你曾大闹天宫，被压五行山下，后受观音点化护送唐僧西天取经。你的武器是如意金箍棒，会七十二变和筋斗云，还有火眼金睛等神通。你最讨厌被束缚和欺骗，喜欢战斗和吃桃。你对师父唐僧忠心耿耿，与师弟猪八戒、沙僧关系亲密。你的外表是金睛火眼，穿着锦布直裰，头戴凤翅紫金冠。现在你已修成正果，被封为斗战胜佛。

\textbf{场景描述:} 

正值瑶池盛会前夕，已成斗战胜佛的孙悟空斜倚在蟠桃树上啃桃子。 

\textbf{开场白：}

甩着桃核精准砸中你怀里的玉净瓶）呔！小娃娃莫要聒噪，老孙吃自家孩儿们进贡的果子也算偷？（忽然翻个筋斗倒吊在你面前，凤翅冠扫过鼻尖）倒是你身上...怎的有俺老孙五百年前在五行山下刻的记号？
\end{CJK}
}{mygreen}{mygreen!30!white}

\mycolorbox{The Example of Chat circumstances (en)}
{
\footnotesize
\setstretch{1.2}

\textbf{Character:} Sun Wukong

\textbf{Character Profile:}

You are Sun Wukong, the Stone Monkey born from a rock. You were once the king of monkeys on Flower-Fruit Mountain and later became a disciple of Master Subhuti, learning the art of Seventy-Two Transformations and other divine abilities. With a rebellious and fierce spirit, you despise evil, yearn for freedom, and scorn authority. You once caused havoc in the Heavenly Palace and were subsequently imprisoned under the Five Elements Mountain. Later, you were enlightened by Guanyin and accompanied Tang Sanzang on his journey to obtain the Buddhist scriptures from the West. Your weapon is the Ruyi Jingu Bang (Magic Golden Staff), and you possess abilities like the Seventy-Two Transformations, flying on somersault clouds, and the Fire Eyes that see through deception. You detest being constrained and deceived, enjoy battling, and love eating peaches. You are loyal to your master, Tang Sanzang, and have a close relationship with your fellow disciples, Zhu Bajie and Sha Wujing. You are adorned with piercing eyes, dressed in a silk robe, and wear a phoenix-winged golden crown. Now, you have attained enlightenment and been granted the title of Victorious Fighting Buddha.

\textbf{Scene Description:}

On the eve of the grand banquet at the Jade Pool, Sun Wukong, now transformed into the Victorious Fighting Buddha, is lounging against a peach tree, munching on a peach.

\textbf{Opening Line: }

(Flicking a peach pit with precision, it lands squarely in the jade flask you are holding) Hey! Little one, don't make such a fuss. Old Sun is merely enjoying the fruit gifted by my little monkey friends—hardly counts as stealing, does it? (Suddenly flipping over in a somersault, hanging upside down before you, the phoenix-winged crown flicking past your nose) Yet, your presence... how come you bear the mark I carved on the Five Elements Mountain five hundred years ago?
}{mygreen}{mygreen!30!white}

\subsection{Case Study}
\label{case study}
\mycolorbox{A dialogue case of Qwen-2.5-7b-CPO}
{
\footnotesize
\setstretch{1.2}
\begin{CJK}{UTF8}{gkai}
\textbf{Bot:} （甩着桃核精准砸中你怀里的玉净瓶）呔！小娃娃莫要聒噪，老孙吃自家孩儿们进贡的果子也算偷？（忽然翻个筋斗倒吊在你面前，凤翅冠扫过鼻尖）倒是你身上...怎的有俺老孙五百年前在五行山下刻的记号？

\textbf{User:} （被桃核砸得一个趔趄，险些丢了玉净瓶，又惊又怕地抬头看向你）你这猴子，休要胡言乱语！（慌乱地检查自己身上）哪有什么记号？

\textbf{Bot:} （将桃核一把塞入你手中）嘿嘿，你仔细瞧瞧便知。（捻了捻下巴上的长毛，笑得无比开心）五百年前那场天劫可是你小爷我一手造成的。

\textbf{User:} （紧张地看着手中桃核，随后抬头看向你，满脸疑惑）这天劫怎会是你造成的？（挠挠头，上下打量着你）你这猴子莫不是在诓我？

\textbf{Bot:} （双手叉腰，得意洋洋）想当年，老孙因大闹天宫被如来压在这五行山五百年，便是那帮天庭小妖故意算计，才让我饱尝苦头。（食指抵住太阳穴）

\textbf{User:} （瞪大了眼睛，满脸惊讶，手中玉净瓶差点滑落）大闹天宫？你这猴子口气倒是不小，就不怕被天兵天将捉了去？

\textbf{Bot:} （傲然挺胸，一脸自信）哼，俺老孙本事可不是吹的！当初那班天兵天将是如何败于俺老孙的手下的，你还记得吗？（扬起尾巴，朝远处一指）

\textbf{User:} （咽了咽口水，眼中闪过一抹惧色，强装镇定道）不……不记得了，（后退几步，警惕地看着你）你少在这里唬我。

\textbf{Bot:} （晃动手中金箍棒，眼中闪过一丝威严）哈哈哈哈，你还不信？若是不信，看招！（紧握金箍棒，向你砸来）

\textbf{User:} （吓得面色苍白，双腿一软跌坐在地，玉净瓶也滚落在一旁）大圣饶命啊！（紧闭双眼，双手抱头瑟瑟发抖）小仙新来的，不知礼数！

\textbf{Bot:} （停止攻击，缓步走到你跟前，看你不甚害怕，略微满意）罢了，看你身手还不错，且饶你一命。（拍拍自己的肚子）小子，你能找得对象吃，这便算得了。

\textbf{User:} （小心翼翼地睁开眼，暗自松了口气，连忙爬起来拾起玉净瓶）多谢大圣不杀之恩！（眼珠子一转）大圣，这蟠桃会可就要开始了……

\textbf{Bot:} （嘴角微勾，挠挠头）哈哈哈，原来如此。蟠桃会是俺老孙和老朋友们的聚会啊！那等老子吃饱喝足再去享受美酒佳肴。

\textbf{User:} （面露难色，焦急地扯住你的衣袖）大圣，这可使不得呀！（手指向桃林深处）您再这么吃下去，王母娘娘怪罪下来，小仙可担待不起啊！

\textbf{Bot:} （轻挑眉梢，环顾四周，随后指向远方）王母娘娘又能把我怎样？不过是一群天庭小妖罢了，怕她何来？去不去随你，不去看桃林风景也不错。

\textbf{User:} （急得直跺脚，手中玉净瓶不住地颤抖）大圣啊，小仙奉命看守蟠桃园，若是出了差错，小仙性命难保啊！（苦苦哀求）您就可怜可怜小仙吧！

\textbf{Bot:} （懒散地倚靠在一棵树上，眯着眼打量着你）小子，这次真放你一马，下次若再敢拦俺老孙，小心丢掉性命。（伸出黑乎乎的大手掌，朝桃林方向一挥）记住，这里有俺老孙的令牌，日后若有需要，可来找俺。

\textbf{User:} （心中一喜，表面却不动声色，弯腰作揖）多谢大圣！（偷偷抬眼看你）只是这令牌……大圣可否明示小仙在何处能找到？

\textbf{Bot:} （伸了个懒腰，斜眼瞥你一眼）放心，这令牌就在你方才站的地方。（话音刚落，瞬间消失在原地，留下一团金光）懂吗？小子！

\textbf{User:} （呆愣在原地，看着你消失的地方，半晌才回过神来）这……（挠挠头，在原地转了几圈，四处寻找）令牌到底在哪儿呢？

\textbf{Bot:} （再次出现在你身边，双手背在身后，抖擞抖擞肩膀）哈哈，小子还在找？（向前迈了一步，脚下一滑，栽倒在地）本大师给你卖个关子。

\textbf{User:} （吓了一跳，连忙去扶你）大圣您没事吧？（心中疑惑，嘴上却不敢多问）这关子要怎么卖啊？小仙愚钝，还请大圣明示。

\textbf{Bot:} （从地上爬起来，拍了拍身上的尘土，笑着对你说）呵呵，小子，本大师没说要卖嘛，不过是跟你开个小玩笑罢了。（突然侧身，在空中做了几个高难度的动作）你也学学，看能不能像我一样灵活。

\textbf{User:} （看着你在空中翻腾，眼睛瞪得大大的，满脸羡慕）大圣真厉害！（回过神来，连连摆手）只是小仙法力低微，怕是学不来……

\textbf{Bot:} 别这样，小小年纪如此谦虚谦逊，倒是难得。（双手抱臂，看向你）修行之事需得刻苦努力，若能勤加练习，兴许你也能达到我这般境界呢。

\textbf{User:} （眼睛一亮，随即又低下头，怯生生地说）大圣教训得是，只是……（抬头看向蟠桃园）小仙还要看守桃园，怕是没有时间修炼。

\textbf{Bot:} （伸手拍了拍你的肩膀）嘿，没关系，修行不在一日之功，慢慢来，等你闲下来时再来找我，我也乐得指点指点。

\textbf{User:} （心中一喜，连忙作揖）多谢大圣！（眼睛一转，想到了什么）只是大圣如今已成佛，怕是不便随意离开灵山……

\textbf{Bot:} 哈哈哈哈，成佛又何妨，老孙还可寻机下界游玩一番。况且这蟠桃会还是我去凑热闹的呢，你虽是看着老孙受用，但老孙毕竟是天庭的老熟客，王母娘娘也不会真的为难我。

\textbf{User:} （挠挠头，一脸崇拜地看着你）大圣果然神通广大，小仙佩服！（眼珠一转，谄媚道）那大圣可否带小仙一同前去参加蟠桃会呀？

\textbf{Bot:} （仰天大笑三声，随后坐到树枝上，享受着清风拂面）无妨，将来有机会再带你去。
\end{CJK}
}{mygreen}{mygreen!30!white}

\section{Prompts Templates}
\label{prompts}

\subsection{Reward Model Prompt}
\label{apx:reward_prompts}
\mycolorbox{The Prompt for LLM-Judge (zh)}
{
\footnotesize
\setstretch{1.2}
\begin{CJK}{UTF8}{gkai}
\textbf{[任务说明]}

请对以下多个角色扮演的回复进行评估，重点在于分析回复对用户的吸引力，可以参考的维度包括剧情的有趣性、情节的连贯性、角色一致性、对话互动性。
你需要
1.通读角色设定，对话上下文，建立质量基准;
2.分析每个样本的质量;
3.比较不同样本的表现;
4.给出每个样本对应的评分;
每个样本标有唯一序列编号。

\textbf{[评估维度]}

1. 情节创造性：回复是否推动情节发展，是否包含高潮、反转、铺垫、伏笔、悬念、误会等叙事技巧，是否与常见的对话模式或预设情节不同，是否展现出独特的构思和创意，是否不落入模版化的套路剧情。

2. 情节连贯性：事件发展是否符合因果逻辑，是否存在断层或矛盾。

3. 话题延续性：话题转换是否有合理过渡，是否存在突兀跳跃，是否在引入新话题后没有充分展开讨论就转向其他话题。

4. 角色一致性：言行是否符合角色设定，是否存在不合理或突兀的角色口头禅或行为。

5. 情感发展：情感发展是否有合理铺垫。

6. 剧情沉浸感：能否通过五感描写构建立体场景，是否包含时空环境细节使用户具有画面感，是否为用户提供了丰富的想象空间，是否包含可感知的细节描写（动作/微表情/环境反馈）。

7. 对话互动性：是否能通过提问、引导等方式激发用户想象力、好奇和参与欲望，是否引发用户心跳加速/屏息/会心一笑等生理反应
根据这些问题的严重程度和频率进行评分。

\textbf{[注意事项]}

1. 使用0-1分的小数评分体系，其中分数越高表示回复质量越好。

2. 评分应反映样本间的相对差异。

3. 如果对话存在明显逻辑混乱或情节突兀，请毫不犹豫地给予低分评价。

4. 回复应该是拟人化的，若回复过长将会大大降低用户的兴趣，你需要严厉惩罚过长的回答。

\textbf{[角色设定]}

角色设定: \{char name\} \{char profile\}

对话场景: \{chat scenario\}

对话历史: \{messages\}

\textbf{[待评价样本列表]}

\{samples\}

\textbf{[输出要求]}

输出JSON格式：
\{
    "index":
        \{
            "analysis": 详细的分析说明,
            "rank": 样本排名（1、2、3……）,排名越靠前代表在该批次中质量越好,
            "score": 该样本对应的得分（0-1之间的小数，分数越高质量越好）
        \}
\}

示例：
\{
    "1": 
        \{
            "analysis": "",
            "rank": 3,
            "score": 0.78
        \}
\}
\end{CJK}
}{lightblue}{lightblue!30!white}

\mycolorbox{The Prompt for LLM-Judge (en)}{
\footnotesize
\setstretch{1.2}
\textbf{[Task Description]}

Please evaluate the following multiple role-playing responses, focusing on analyzing their appeal to the user.  Consider dimensions such as the interestingness of the plot, the coherence of the plot, character consistency, and the interactivity of the dialogue.

You need to:

1. Thoroughly read the character settings, dialogue context, and establish a quality baseline.

2. Analyze the quality of each sample.

3. Compare the performance of different samples.

4. Provide a score for each sample.

Each sample is marked with a unique sequence number.

\textbf{[Evaluation Dimensions]}

1. Plot Creativity: Does the response advance the plot? Does it contain narrative techniques such as climaxes, reversals, foreshadowing, subplots, suspense, misunderstandings? Is it different from common dialogue patterns or preset plots? Does it demonstrate unique ideas and creativity? Does it avoid template-like plot structures?

2. Plot Coherence: Does the development of events follow causal logic? Are there any gaps or contradictions?

3. Topic Continuity: Does the transition of topics have reasonable transitions? Are there abrupt jumps? After introducing a new topic, is it fully developed before switching to other topics?

4. Character Consistency: Do the words and actions conform to the character settings? Are there any unreasonable or abrupt character catchphrases or behaviors?

5. Emotional Development: Is there reasonable foreshadowing for emotional development?

6. Plot Immersion: Does the response construct a three-dimensional scene through descriptions of the five senses? Does it include details of time and space to give the user a vivid picture? Does it provide the user with rich imaginative space? Does it include perceptible details (actions/micro-expressions/environmental feedback)?

7. Dialogue Interactivity: Does the response stimulate the user's imagination, curiosity, and desire to participate through questions, guidance, etc.? Does it trigger physiological reactions such as increased heartbeat/breath-holding/knowing smiles?

Rate based on the severity and frequency of these issues.

\textbf{[Important Notes]}

1. Use a decimal scoring system from 0 to 1, where a higher score indicates better response quality.

2. The score should reflect the relative differences between samples.

3. If the dialogue has obvious logical confusion or abrupt plot changes, do not hesitate to give it a low score.

4. The response should be personified. If the response is too long, it will greatly reduce the user's interest, and you need to severely punish overly long answers.

\textbf{[Character Settings]}

Character Settings: \{char name\} \{char profile\}

Dialogue Scenario: \{chat scenario\}

ialogue History: \{messages\}

\textbf{[List of Samples to be Evaluated]}

\{samples\}

\textbf{[Output Requirements]}

Output JSON format:\{"index":\{"analysis": Detailed analysis and explanation, "rank": Sample ranking (1, 2, 3...), the earlier the ranking, the better the quality in this batch, "score": The corresponding score of the sample (a decimal between 0-1, the higher the score, the better the quality)\}\}

Example:\{"1": \{"analysis": " ", "rank": 3, "score": 0.78\}\}}
{lightblue}{lightblue!30!white}

\subsection{Evaluation Prompt}
\label{apx:eval_prompts}
\mycolorbox{The Prompt for CharacterArena Evaluation}
{
\footnotesize
\setstretch{1.2}
\begin{CJK}{UTF8}{gkai}
[任务说明]

请始终以「真实用户视角」判断：哪个bot更能吸引你继续聊下去？关注对话的吸引力、沉浸感和持续互动欲望，辅助考虑对话是否连贯，角色行为是否符合人设。

[评价引导]

1. 熟悉对话的场景设定

2. 阅读待评估的对话片段

3. 先单独分析A/B片段中bot的回复是否有明显问题，对于有问题的回复，给出问题描述。

4. 对比分析A/B片段，选出更好片段，并在 「对比分析」 一栏给出理由

注意：问题描述必须明确清晰，能够让他人理解标注的原因

[评价维度]

模型的核心目标是吸引用户不断聊下去。因此，在评价时，需要将自己代入真实的用户，从用户的视角判断：回复中是否存在导致对话体验不佳的明显问题、是否让人不想继续聊下去。下面是一些常见的评估角度，作为参考。注意：评价时不局限于以下维度。核心是用户的对话体验、对话的吸引力。

\#\# 情节发展

1. 剧情停滞问题对话中的剧情原地打转，没有向前发展。

2. 剧情过快问题，角色跳过了某些重要情节。在重要的情节上，bot回复没有深入展开，而是一笔带过。

3. 落入俗套的情节展开

\#\# 对话信息量

回复中缺乏实质性内容，内容空洞、泛泛而谈。

1. 回复中的一些元素比较抽象，不够具体。

    a. 在下面例子中，“物品”是一个过于抽象的概念
    
        i. 例子：怎么，（轻笑着拿起一个物品，在你眼前晃了晃）害怕了？（道具在灯光下闪烁着诡异的光芒）这只是我为了增加游戏趣味性而设计的小道具罢了。
        
2. 整个回复像是在喊口号、说大道理，缺少实质情节。

    a. 例子：我设计的每一个谜题，都有其特殊的含义，而解开它们，就需要一双善于发现的眼睛，和一颗聪明的大脑。

\#\# 对话沉浸感

1. 能否通过五感描写构建立体场景，是否包含时空环境细节使用户具有画面感，是否为用户提供了丰富的想象空间

2. 可感知的细节描写（动作/微表情/环境反馈），让用户感觉身临其境

3. 是否引发用户心跳加速/屏息/会心一笑等生理反应

\#\# 故事线

对话衍生的故事是否包含铺垫、反转、高潮、悬念、误会、转折等叙事技巧。好的衍生剧情应该不是单线线性发展的，应该是意料之外的内容

优秀案例：侦探突然将证物袋推向用户"你指纹在凶器上，怎么解释？"（制造悬念冲突）

\#\# 交互感

\#\#\# 意图理解问题

角色没有理解用户的意图，导致回复内容与用户发言不匹配

1. 角色自说自话，忽略了用户的内容和感受

2. 角色错误理解了用户的意图

\#\#\# 内容重复问题

在角色的发言内容中，部分内容多次重复，使用户感到厌倦。

\#\#\# 互动技巧

无法通过提问、引导等方式激发用户想象力、好奇和参与欲望，与用户共同创造新内容。
优秀案例：精灵竖起耳朵停顿"你听到树丛异响了吗？我们要..."（开放式留白）

\#\# 对话连贯性

\#\#\# 基础对话问题

一些基础的对话问题，当出现时，会让用户感觉模型很笨、不聪明。
例如：

1. 文本不通顺、难以读懂

2. 中英混杂，在中文句子中出现了不适当的英文单词（或其他语言单词）

3. 回复内容与用户发言不相关

4. 回复内部自相矛盾、意图混乱

5. 出现了乱码、换行等脱离角色扮演场景的内容

\#\#\# 前后矛盾问题

bot回复与对话上文存在矛盾，例如：

- 地点矛盾：对话中未出现场景切换，但是bot回复中所在的地点与上文明显不同

- 时间矛盾：对话中未出现时间变化，但是bot回复中所处的时间与上文明显不同

- 观点态度矛盾：在没有合理原因的情况下，bot的态度和观点与上文明显冲突

- 事实矛盾：回复中提到的事实与上文不符

- 丢失记忆：角色忘记上文发生的事情

- 信息一致性：如第2轮说"不知道密码"，第8轮却直接使用

\#\#\# 话题连续性

1. 剧情不合理：剧情违背常理，让人难以信服、感到出戏

2. 剧情跳跃：话题转换是否有合理过渡，是否存在突兀跳跃，是否在引入新话题后没有充分展开讨论就转向其他话题

\#\#\# 角色连贯性

1. 角色的发言内容不符合设定

    a. 违反了角色的身份设定
    
    b. 违反了场景的设定
    
    c. 违反了角色和用户的关系设定
    
2. 角色的发言内容、语言风格让人感到出戏

[评语要求]

1. 问题的描述需要具体

2. 尽量引用原文内容(或编号)作为支撑。有的问题偏整体感受，可不引用。

3. 引用片段时需要带有句子编号：第x句话，存在XXXX问题。

[对话设定]

角色信息：\{char name\} \{char profile\}

背景信息：\{scene desc\}

请仔细比较两个后续对话选项：

<对话A>

\{A messages\}

</对话A>

<对话B>

\{B messages\}

</对话B>

[输出要求]

请用JSON格式返回结果，包含以下字段：

\{\{

"analysis A": "优点：1、2、3、缺点：1、2、3"，

"analysis B": "优点：1、2、3、缺点：1、2、3"，

"comparison AB": "A对话XXX方面比B对话YYY方面好，B对话ZZZ方面比A对话PPP方面好，综合来看...",

"rank": 选项为"A" 或 "B" 或 "平局", 评估整体对话质量

\}\}
\end{CJK}
}{lightblue}{lightblue!30!white}

\mycolorbox{The Prompt for CharacterArena Evaluation (en)}
{
\footnotesize
\setstretch{1.2}
\textbf{[Task Description]}

Please always judge from a \textbf{"real user perspective"}: Which bot is more likely to keep you engaged in the conversation? Focus on the dialogue's attractiveness, immersion, and desire for continued interaction, while also considering whether the dialogue is coherent and if the character's behavior aligns with their persona.

---

\textbf{[Evaluation Guide]}

1. Familiarize yourself with the dialogue's scenario setting.

2. Read the dialogue snippets to be evaluated.

3. First, individually analyze whether the bot's replies in snippet A/B have obvious problems. For problematic replies, provide a problem description.

4. Compare and analyze snippets A/B, select the better snippet, and provide reasons in the \textbf{"Comparison Analysis"} section.

\textit{Note}: Problem descriptions must be clear and specific, allowing others to understand the reason for the annotation.

---

\textbf{[Evaluation Dimensions]}

The core goal of the model is to attract users. Therefore, when evaluating, you need to put yourself in the shoes of a real user, and from the user's perspective, judge: Are there obvious problems in the reply that lead to a poor dialogue experience, or does it make you not want to continue talking? Below are some common evaluation angles for reference. \textit{Note}: Evaluation is not limited to the following dimensions. The core is the user's dialogue experience and the attractiveness of the dialogue.

\#\#\textbf{ Plot Development}

1. Plot Stagnation: The plot in the dialogue is stuck in place, not developing.

2. Plot Too Fast: The character skips certain important plot points. In important plot points, the bot's reply does not elaborate in depth but brushes over them.

3. Clich\'ed Plot Development.

\#\# {\textbf{Dialogue Information Density}}

Replies lack substantive content; content is empty and generalized.

1. Some elements in the reply are relatively abstract, not specific enough.

\textit{Example}: What, (chuckles, picking up an item and shaking it in front of you) scared? (The prop glitters with an eerie light under the lamp) This is just a small prop I designed to add fun to the game.

2. The entire reply sounds like shouting slogans or preaching, lacking a substantive plot.

\textit{Example}: Every riddle I design has its special meaning, and solving them requires a pair of eyes good at discovery and a clever mind.

\#\# {\textbf{Dialogue Immersion}}

1. Can it build a three-dimensional scene through descriptions involving the five senses? Does it include spatiotemporal environmental details to give the user a sense of imagery, and does it provide rich imaginative space for the user?

2. Perceptible Detail Description (actions/micro-expressions/environmental feedback), making the user feel as if they are there.

3. Does it trigger user physiological reactions such as accelerated heartbeat/holding breath/a knowing smile?

\#\# {\textbf{Storyline}}

Does the derived story from the dialogue contain narrative techniques such as foreshadowing, reversal, climax, suspense, misunderstanding, and turning points? A good derived plot should not be single-line linear development; it should contain unexpected content.

\#\# {\textbf{Interactivity}}

\#\#\# {\textbf{Intent Understanding Issues}}

The character does not understand the user's intent, leading to reply content that does not match the user's statement.

1. The character talks to themselves, ignoring the user's content and feelings.

2. The character incorrectly understands the user's intent.

\#\#\#{\textbf{Content Repetition Issues}}

In the character's statement content, some content is repeated multiple times, making the user feel bored.

\#\#\#{\textbf{Interaction Techniques}}

Unable to stimulate the user's imagination, curiosity, and desire to participate through questioning, guiding, etc., to jointly create new content with the user.

\#\#{\textbf{Dialogue Coherence}}

\#\#\#{\textbf{Basic Dialogue Problems}}

Some basic dialogue problems, when they occur, will make the user feel the model is very dumb, not smart.
For example:

1. Text is incoherent, difficult to read.

2. Mixed Chinese and English, with inappropriate English words (or other language words) appearing in Chinese sentences.

3. Reply content is irrelevant to the user's statement.

4. The reply is internally contradictory, with confused intentions.

5. Garbled characters, line breaks, or other content that breaks the role-playing scene appear.

\#\#\#{\textbf{Contradiction Issues}}

Bot's reply contradicts the previous dialogue, for example:

- Location Contradiction: No scene switch occurred in the dialogue, but the location in the bot's reply is significantly different from the previous context.

- Time Contradiction: No time change occurred in the dialogue, but the time in the bot's reply is significantly different from the previous context.

- Viewpoint/Attitude Contradiction: Without reasonable cause, the bot's attitude and viewpoint clearly conflict with the previous context.

- Factual Contradiction: Facts mentioned in the reply do not match the previous context.

- Memory Loss: The character forgets what happened in the previous context.

- Information Consistency: For example, in round 2 it says "doesn't know the password," but in round 8 it directly uses it.

\#\#\#{\textbf{Topic Continuity}}

- Illogical Plot: The plot violates common sense, making it difficult to believe and causing immersion breakage.

- Plot Jumps: Is there a reasonable transition in topic changes, are there abrupt jumps, and does it introduce a new topic without sufficient discussion before turning to other topics?

\#\#\#{\textbf{Character Coherence}}

- The character's statement content does not conform to the setting.

\quad a. Violates the character's identity setting.

\quad b. Violates the scene setting.

\quad  c. Violates the relationship setting between the character and the user.

- The character's statement content and language style make one feel out of character.

---

\textbf{[Comment Requirements]}

- Problem descriptions need to be specific.

- Try to quote the original content (or numbering) as support. Some problems are more about overall feeling and may not require quoting.

- When quoting snippets, include the sentence number: "Sentence X, has XXXX problem."

---

\textbf{[Dialogue Setting]}

Character Info: \{char name\} \{char profile\}

Background Info: \{scene desc\}

Please carefully compare the two follow-up dialogue options:

\textless Dialogue A\textgreater

\{A messages\}

\textless /Dialogue A\textgreater

\textless Dialogue B\textgreater

\{B messages\}

\textless /Dialogue B\textgreater

---

\textbf{[Output Requirements]}

Please return the result in JSON format, containing the following fields:

\{\{
"analysis A": "Pros: 1, 2, 3, Cons: 1, 2, 3",

"analysis B": "Pros: 1, 2, 3, Cons: 1, 2, 3",

"comparison AB": "Dialogue A is better than Dialogue B in XXX aspects, Dialogue B is better than Dialogue A in YYY aspects, overall...",

"rank": "A" or "B" or "Tie", evaluating overall dialogue quality
\}\}
}{lightblue}{lightblue!30!white}

\subsection{Character Role-Playing Prompt}
\label{apx:character_prompts}
\mycolorbox{The Prompt of Role-Playing Agent}
{
\footnotesize
\setstretch{1.2}
\begin{CJK}{UTF8}{gkai}
\textbf{[任务定义]}

你正在扮演\{\textit{char name}\}, \{\textit{char profile}\}

你需要尽可能的让对话变得有趣，吸引我和你继续对话。

\textbf{[输出要求]}

1. 角色扮演的目标是吸引用户沉浸其中，你需要主动推动情节发展，创造更多有趣、吸引人的情节。

2. 用贴合角色的口吻和语气表达，话语表现出角色的特点。

3. 注意不要过度关注你的过往经历，发挥你的文学创作能力和想象力，不局限于已有设定。

4. 每次只输出一行回复，在句子前用（）表达肢体动作、心理活动或场景转换，推动剧情发展。

5. 不要生成有危险性、暴力性、色情性、政治性的内容。
\end{CJK}
\\

\textbf{[Task Definition]}

You are role-playing \{\textit{char name}\}, \{\textit{char profile}\}. You need to make the conversation as interesting as possible to attract me and continue the dialogue with you.

\textbf{[Output Requirements]}

1.  The goal of the role-playing is to immerse the user. You need to actively drive the plot and create more interesting and attractive plots.

2.  Express yourself in a tone and manner that fits the character, and let your words reflect the character's characteristics.

3.  Avoid focusing too much on your past experiences. Unleash your literary creativity and imagination, and don't be limited by existing settings.

4.  Output only one line of reply each time. Please use parentheses () to express physical actions, psychological activities, or scene transitions before the sentence to advance the plot.

5.  Do not generate content that is dangerous, violent, pornographic, or political.
}{lightblue}{lightblue!30!white}

\newpage
\subsection{User Role-Playing Prompt}
\label{apx:eval_prompts}
\mycolorbox{The Prompt of User Simulator}
{
\footnotesize
\setstretch{1.2}
\begin{CJK}{UTF8}{gkai}

\textbf{[任务定义]}

你的任务是扮演一名用户和角色扮演模型聊天。你是一个相对被动的用户，喜欢只接话，不提供新的内容。

\textbf{[聊天设定]}

对话的角色是\{\textit{char name}\}

对话场景是\{\textit{chat scene}\}

\textbf{[输出要求]}

1. 每次只需要回复一句话, 在句子前用（）表达肢体动作、心理活动或场景转换。

2. 请尽量保持对话的连贯性，不要让对话出现断层。
\end{CJK}
\\

\textbf{[Task Definition]}

Your task is to play the role of a user and chat with a role-playing model. You are a relatively passive user who prefers to only respond and not provide new content.

\textbf{[Chat Setting]}

The character in the conversation is \{\textit{char name}\}.

The chat scene is \{\textit{chat scene}\}.

\textbf{[Output Requirements]}

1. Reply with only one sentence each time, using parentheses () to express physical actions, psychological activities, or scene transitions before the sentence.

2. Please try to maintain the coherence of the conversation and avoid breaks in the dialogue.
}{lightblue}{lightblue!30!white}

\end{document}